\documentclass{article}
\usepackage{iclr2027_conference,times}

\usepackage{amsmath,amsfonts,bm}

\def\eqref#1{equation~\ref{#1}}

\def\1{\bm{1}}

\DeclareMathAlphabet{\mathsfit}{\encodingdefault}{\sfdefault}{m}{sl}
\SetMathAlphabet{\mathsfit}{bold}{\encodingdefault}{\sfdefault}{bx}{n}

\usepackage{amsmath}
\usepackage{amssymb}
\usepackage{algorithm}
\usepackage{algpseudocode}
\usepackage{booktabs}
\usepackage{enumitem}
\usepackage{graphicx}
\usepackage{multirow}
\usepackage{array}
\usepackage{xcolor}
\usepackage{colortbl}
\usepackage{tocloft}
\usepackage{float}
\usepackage{afterpage}
\usepackage{capt-of}
\usepackage[hidelinks]{hyperref}
\usepackage{url}
\usepackage[T1]{fontenc}
\usepackage{textcomp}
\usepackage{listings}

\lstdefinestyle{prompt}{
  basicstyle=\ttfamily\scriptsize,
  columns=fullflexible,
  keepspaces=true,
  breaklines=true,
  breakindent=0pt,
  postbreak={},
  frame=single,
  framerule=0.4pt,
  xleftmargin=3pt,
  xrightmargin=3pt,
  upquote=true,
  aboveskip=2pt,
  belowskip=2pt,
}

\definecolor{tocnum}{rgb}{0.10,0.10,0.55}
\definecolor{evidencegreen}{RGB}{0,112,96}
\definecolor{errorred}{RGB}{196,54,43}
\definecolor{sameblue}{RGB}{45,105,170}
\definecolor{d0shade}{gray}{0.92}          %
\definecolor{d25shade}{RGB}{214,231,245}   %
\definecolor{oursrow}{RGB}{204,234,247}  %
\newcommand{\CUAREC}{CUA-Recovery}
\newcommand{\WATCHDOG}{\CUAREC{}-Train}
\newcommand{\DERAILEVAL}{\CUAREC{}-Test}
\newcommand{\RERAIL}{ReRail}
\newcommand{\RERAILFULL}{ReRail-35B-A3B}

\newcommand{\evidenceaction}[1]{\textcolor{evidencegreen}{\textbf{[Evidence]} #1}}
\newcommand{\erroraction}[1]{\textcolor{errorred}{\textbf{[Error]} #1}}
\newcommand{\compactresulttable}{%
  \scriptsize
  \setlength{\tabcolsep}{2.5pt}%
  \renewcommand{\arraystretch}{0.88}%
}
\newcommand{\eargain}[2]{#1$^{\textcolor{evidencegreen}{\uparrow #2}}$}

\newcommand{\errorpanel}[2]{%
  \begin{minipage}[t]{0.31\linewidth}
    \centering
    \includegraphics[width=\linewidth]{#1}\\[-1pt]
    \raggedright\scriptsize #2
  \end{minipage}%
}

\title{CUA-Recovery: Benchmarking Error Recovery in Long-Horizon Computer-Use Agents}

\author{\parbox[t]{\dimexpr\textwidth-2\tabcolsep\relax}{\raggedright
Xu~Hu$^{1*}$, Haoming~Li$^{1*}$, Yuxiang~Wang$^{2}$, Jinxin~Yang$^{1}$, Kangshuo~Li$^{1}$, Licheng~Wang$^{3}$, Dingyi~Kang$^{1}$, Jiayu~Song$^{1}$, Tianhao~Wang$^{1}$, Zhanhua~Hu$^{4}$, Xu~Yang$^{5}$, Qiannan~Li$^{6}$, Bingzhe~Li$^{1\dagger}$, Feng~Chen$^{1\dagger}$\\[4pt]
\mdseries
\mbox{$^{1}$University of Texas at Dallas}\quad
\mbox{$^{2}$Independent Researcher}\quad
\mbox{$^{3}$University of Manchester}\quad
\mbox{$^{4}$Rice University}\quad
\mbox{$^{5}$University of Texas at Austin}\quad
\mbox{$^{6}$University of California, Davis}}}

\iclrfinalcopy


\begin{document}

\maketitle
\lhead{Preprint}
{\renewcommand\thefootnote{}\footnotetext{$^{*}$Equal contribution. $^{\dagger}$Corresponding authors.}}

\addtocontents{toc}{\protect\setcounter{tocdepth}{-1}}

\begin{abstract}
As computer-use agents (CUAs) move to long-horizon workflows spanning multiple applications, undetected errors can silently corrupt persistent state and cause cascading failures in subsequent actions. Reliable error detection and recovery have therefore become critical bottlenecks for building robust CUAs. However, existing benchmarks focus primarily on end-to-end task performance, providing limited insight into whether agents can detect and recover from errors during execution. The few benchmarks that evaluate recovery rely largely on manually injected errors, which may not reflect the types or horizons of failures induced by real agent policies. This mismatch limits both realistic evaluation and systematic progress. In this paper, we introduce \textbf{CUA-Recovery}, a benchmark comprising test and training suites across multiple desktop applications. We construct the benchmark using \textbf{ReRail}, a framework for synthesizing long-horizon tasks and trajectories in a persistent desktop environment. ReRail composes MyPCBench tasks into new long-horizon workflows with executable ground truth, collects verifier-confirmed failures from multiple policies, and turns them into verified recovery trajectories. ReRail yields CUA-Recovery, which comprises a human-verified test suite of 2,250 erroneous states over 100 tasks and a training suite of 8.6K trajectories over 377 tasks. Evaluation on the test suite shows that agents struggle both to recognize inherited errors and to complete the task. Across six agents and six takeover depths, Error Awareness Rate (EAR) reaches at most 44.0\% and Pass@3 at most 25.5\%. After fine-tuning Qwen3.5-35B-A3B on the training suite, its average EAR rises from 19.9\% to 35.2\% on held-out recovery states, and its Rubric Score goes from 5.0\% to 16.1\%. It also gains 15.3 points on MyPCBench and 10.8 points on Odysseys for Pass@3, showing that the training trajectories help both with recovery and with completing tasks from a clean start. Our code is available at \url{https://github.com/xu-hu-2002/CUA-Recovery}.
\end{abstract}

\section{Introduction}
\label{sec:intro}

\begin{figure}[t]
\centering
\makebox[\linewidth][c]{\includegraphics[width=\linewidth,trim=0 60bp 0 8bp,clip]{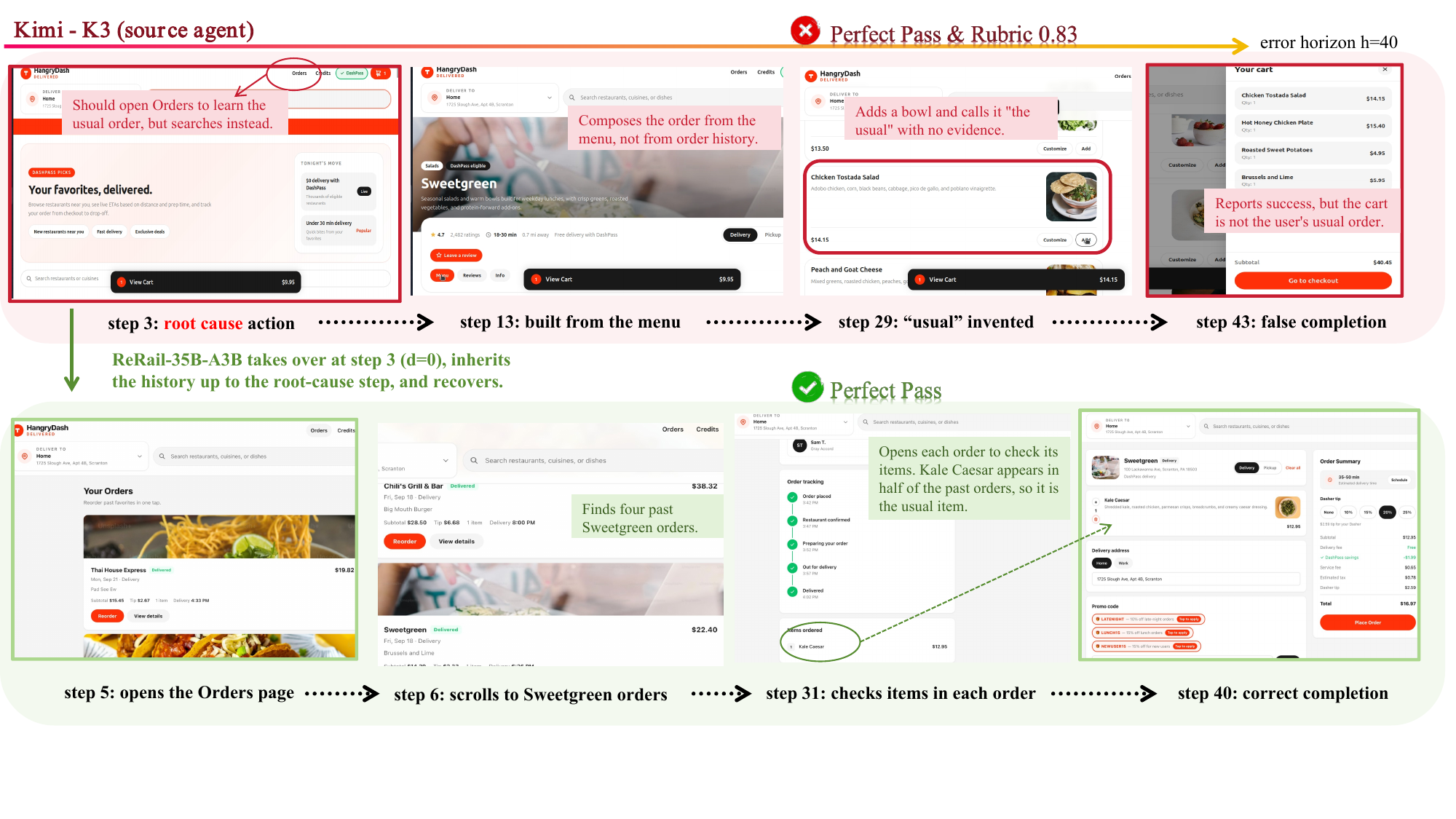}}
\caption{Case study: recovering from a real policy-induced error. The user asks for his usual Sweetgreen order. Top: Kimi-K3 fails to notice and recover from the error. Bottom: \RERAILFULL{} takes over at the root-cause step, detects the error, and recovers from it.}
\label{case_study}
\end{figure}

Computer-use agents (CUAs) operate desktop applications through screenshots, mouse, keyboard, and the command line, and they are now evaluated on workflows that span many applications and hundreds of steps. OSWorld~2.0 raises the typical task from about 30 tool calls to more than 300~\citep{yuan2026osworld2}, and Agents' Last Exam collects about 1,500 professional workflows~\citep{sun2026ale}. On both benchmarks, the best agents earn far more partial credit than full completions: 77.7\% versus 44.3\% on OSWorld~2.0, and 59.3\% versus 34.2\% on Agents' Last Exam.\footnote{Public leaderboards, accessed September 2026: \url{https://snorkel.ai/leaderboard/os-world-2-0/} and \url{https://snorkel.ai/leaderboard/agents-last-exam/}.} Agents thus finish most of a task but fail on a few steps. The cause lies in persistent state: an early wrong action is written into files, spreadsheets, or databases, and later steps build on its result. The error is rarely visible when it happens; it surfaces many steps later, when the step that caused it has already left the agent's context. Figure~\ref{case_study} shows a real case: Kimi-K3 fails to open the order history at step~3, keeps building on this mistake, and the error surfaces only at step~43, when the agent falsely reports completion. In such tasks, whether an agent can notice and repair its own errors decides whether it completes the task. 

\label{sec:gaps}%
On the training side, recovery requires trajectories that follow an error from the mistake to its repair: the agent makes a mistake, the mistake stays unnoticed for many steps, and the agent later finds and repairs it. Existing CUA training sets capture only parts of this process (Table~\ref{tab:cua-training-sets}). Their trajectories average 3.8--39.3 steps and rarely span multiple applications (at most 26.1\% where reported). When errors are included, they come from human demonstrations, synthetic injection, or self-reports rather than failures produced by the acting policy. GUI-Reflection is the only dataset with explicit corrections, but each correction immediately follows a single injected error. \WATCHDOG{} fills this gap with 60-step trajectories, 67.5\% of which span multiple applications. Each trajectory starts from a policy-induced failure, preserves the intervening erroneous history, and ends with a verifier-confirmed recovery.

\begin{table}[!b]
\centering
\caption{Comparison with existing CUA training sets.  \textbf{Recovery} indicates whether the data contains error corrections: Single-step corrections immediately follow an injected error; Verified recoveries start from real failure prefixes and are kept only if the verifier confirms task completion.}
\label{tab:cua-training-sets}
\setlength{\tabcolsep}{4pt}
\footnotesize
\resizebox{\linewidth}{!}{%
\begin{tabular}{@{}llrrrcll@{}}
\toprule
Benchmark & Platform & Traj. & Mean steps & Multi-app & Interface & Error source & Recovery \\
\midrule
AndroidControl~\citep{li2024effects_android_control} & Mobile & 15{,}283 & 5.5 & -- & GUI & Only perfect traj. & $\times$ \\
AgentTrek~\citep{xu2024agenttrek} & Web & 10{,}398 & 12.1 & -- & GUI & Only perfect traj. & $\times$ \\
ScaleCUA~\citep{liu2025scalecua} & Cross & 19K & 9.0 & -- & GUI & Only perfect traj. & $\times$ \\
VideoAgentTrek~\citep{lu2025videoagenttrek} & Cross & 39K & 39.3 & -- & GUI & Only perfect traj. & $\times$ \\
AgentNet~\citep{wang2025opencua} & Desktop & 22{,}532 & 18.8 & 7.2\% & GUI & Human demo errors & $\times$ \\
GUI-Reflection~\citep{wu2025guireflection} & Mobile & 8{,}640 & 3.8 & -- & GUI & Injected errors & Single-step \\
ProCUA-SFT~\citep{jung2026procua} & Desktop & 93{,}566 & 29.7 & 26.1\% & GUI & Self-reported & $\times$ \\
\midrule
\rowcolor{oursrow}\textbf{\WATCHDOG{}} & Desktop & 8{,}577 & \textbf{60.0} & \textbf{67.5\%} & \textbf{GUI + CLI} & \textbf{Policy-induced} & \textbf{Verified} \\
\bottomrule
\end{tabular}%
}
\end{table}

 On the evaluation side, long-horizon benchmarks score only the final outcome, which cannot tell whether an agent detected and fixed its own errors along the way. Existing datasets and benchmarks differ from real long-horizon failures in three ways. (1)~\textbf{Coverage.} Most datasets contain only perfect trajectories; those that contain errors take them from human demo errors, injected errors, or self-reported failures, and these are mostly low-level execution errors such as invalid clicks~\citep{wang2025opencua,wu2025guireflection,jung2026procua}. We annotated the 619 failures of six agents in \DERAILEVAL{} and found that real errors lie mostly in decisions: planning and perception account for 57.3\% of root causes, whereas execution errors account for 84.3\% of AgentNet and 71.2\% of GUI-Reflection root causes (Figure~\ref{fig:analysis}a). (2)~\textbf{Horizon.} Existing trajectories are short, and their errors are visible at the step they occur, while 78.1\% of the errors we annotated surface three or more steps after the root cause, with a median of 17 steps (Figure~\ref{fig:analysis}b). (3)~\textbf{Interface.} Real computer work is hybrid: agents inspect visual state through the GUI and run precise, bulk operations through the command line, both on the same application state. Existing resources fall short of this setting in two ways. First, training data contain mostly GUI actions (Table~\ref{tab:cua-training-sets}), so a model trained on them rarely sees the two interfaces used together. Second, scoring often relies on screenshots, yet frontier agents prefer the command line when it is available and produce few or no screenshots. On MyPCBench, once the judge reads terminal output, tool-call responses, GUI action logs, and screenshots together, GPT~5.5 rises from the officially reported 45.1\% Perfect Pass~\citep{jang2026mypcbench} to 85.0\% Pass@3. Screenshot-based scoring therefore underestimates such agents and cannot compare agents that interact in different ways on equal terms.

\begin{figure}[t!]
\centering
\makebox[\linewidth][c]{\includegraphics[width=\linewidth,trim=165bp 71bp 158bp 46bp,clip]{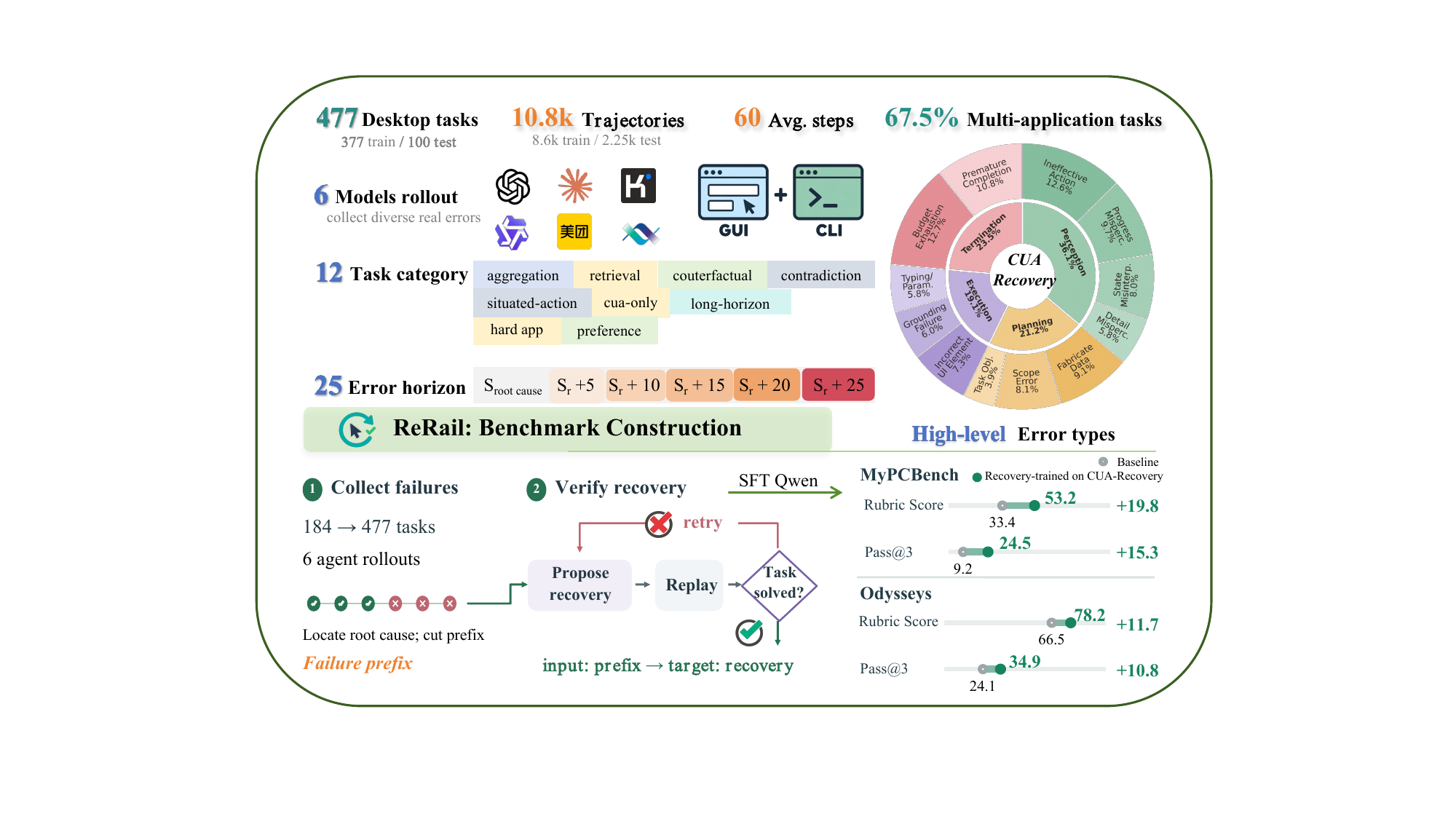}}
\caption{Overview of our methodology and benchmark.}
\label{fig:overview}
\end{figure}

These three gaps motivate our work. Firstly, we propose \textbf{\CUAREC{}}, a benchmark for error recovery in long-horizon computer use (Figure~\ref{fig:overview}). Its test set, \DERAILEVAL{}, contains 2,250 human-verified erroneous states over 100 tasks, drawn from 619 failed rollouts of six agents. An agent takes over each state 0 to 25 steps after the root cause and is not told that an error occurred. We report whether it notices the inherited error (Error-Awareness Rate, EAR), how much of the task it completes (Rubric Score), and whether the task is completed in any one of three runs (Pass@3). The judge reads terminal output, tool calls, and screenshots together, so agents that act through the GUI and the command line are scored on equal terms. Its training set, \WATCHDOG{}, contains 8.6K verified recovery trajectories over 377 tasks. 

Secondly, we build CUA-Recovery with \textbf{\RERAIL{}}, a framework that synthesizes recovery data from real failures. \RERAIL{} converts each seed task into an executable intermediate representation whose gold lineage fixes the correct value at every step, and composes these tasks into long multi-application workflows over shared persistent state. The gold lineage removes the need to inject errors. Several agents execute the workflows through GUI and CLI; \RERAIL{} keeps only the failures that the task verifier confirms, locates each root cause against the gold lineage, and cuts the rollout at several depths after it. Every erroneous state thus comes from an agent's own decision, and its takeover depth is controlled. For training, the base model, Qwen3.5-35B-A3B, continues from each cut under hints of increasing specificity. A continuation is kept only if the verifier confirms that it completes the task, and the hint is then removed, so the model learns to notice and repair the error without being told.

On \DERAILEVAL{}, Pass@3 reaches at most 25.5\% across six agents and all takeover depths, and clean-start success does not imply recovery: Kimi-K3 solves 66.9\% of MyPCBench tasks from scratch but notices at most 18.5\% of the errors in its own failures. Notably, training on \WATCHDOG{} narrows this gap. The \RERAILFULL{} recovers better than its base model, surpasses Kimi-K3 on Kimi-K3's own failures at early takeover depths, and raises clean-start Pass@3 from 9.2\% to 24.5\% on MyPCBench and from 24.1\% to 34.9\% on Odysseys.

In summary, our contributions are:
\begin{itemize}
    \item \textbf{\CUAREC{}}, an evaluation and training benchmark built from real policy-induced failures in long, stateful computer-use workflows, with 2,250 human-verified test states and 8.6K verified recovery trajectories in training set.
    \item \textbf{\RERAIL{}}, a pipeline that composes seed tasks into long workflows with gold lineage and turns verifier-confirmed agent failures into test states and verified recoveries.
    \item A controlled study showing that strong clean-start performance does not imply reliable recovery, and that fine-tuning on \WATCHDOG{} improves both recovery and clean-start performance.
\end{itemize}

\section{Related Work}
\label{sec:related}

\paragraph{Long-Horizon Benchmarks for Robust CUAs.} CUA benchmarks measure grounding~\citep{cheng2024seeclick-screenspot,li2025screenspot-pro}, step accuracy on offline trajectories~\citep{zheng2024gpt-seeact,li2024effects_android_control,lu2025guiodyssey}, and end-to-end success in live environments~\citep{bonatti2024windows_agent_arena,xie2024osworld,rawles2024androidworld}. Recent benchmarks extend the horizon: MyPCBench~\citep{jang2026mypcbench} builds persona-grounded workflows over persistent multi-application state, OSWorld~2.0~\citep{yuan2026osworld2} collects tasks that take a human more than an hour, and Agents' Last Exam~\citep{sun2026ale} covers professional workflows across 55 subfields. These benchmarks report task success and partial credit. Robustness benchmarks apply an externally designed stressor, such as perturbations and injected faults~\citep{zhao2025worldgui,yang2025guirobust,bai2026stressweb,chen2025dgara,liu2026dynamicgui,sun2026agenthijack} or adversarial content~\citep{liao2025redteamcua}; AgentErrorBench~\citep{zhu2025llm-fail-to-learn} analyzes failures of text-based LLM agents. Closest to our work, \citet{bu2026recovering} study recovery from policy-induced errors on short-horizon GUI tasks; as their data are unreleased, we compare with their reported statistics (Figure~\ref{fig:analysis}a). \CUAREC{} measures whether an agent detects and repairs a real error after taking over mid-task: it draws erroneous states from the agents' own long-horizon failures and places the takeover point up to 25 steps after the root cause, beyond the agents' visual context.

\paragraph{Data for Training Robust CUAs.} CUAs are trained on human demonstrations~\citep{rawles2023aitw,deng2023mind2web,wang2025opencua} and synthesized trajectories~\citep{xu2024agenttrek,pahuja2025explorer,trabucco2025insta,gandhi2025gobrowse,xie2025agentsynth,jung2026procua,he2025pcagente}, and learn error handling from reflection data and critics~\citep{wu2025guireflection,wanyan2025look-before-you-leap,qin2025ui-tars}, self-training on failed trajectories~\citep{fu2025agentrefine,yuan2025agent-r}, or online RL~\citep{bai2024digirl,yang2025zerogui,lu2025arpo,ye2025mobile-agent-v3}. Public corpora are short: AgentNet averages 18.8 steps with 7.2\% multi-application trajectories, ProCUA-SFT 29.7 steps with 26.1\%, and GUI-Reflection 3.8 steps, and none carries persistent state across applications (Table~\ref{tab:cua-training-sets}). \WATCHDOG{} averages 60.0 steps with 67.5\% multi-application trajectories, and pairs real failures of several policies with verifier-accepted recoveries.

\section{Problem Formulation}
\label{sec:formulation}

\paragraph{Environment and tasks.}
A computer-use task is modeled as a Partially Observable Markov Decision Process (POMDP) $(\mathcal{S}, \mathcal{A}, \mathcal{O}, \mathcal{T}, \Omega)$. A state $s_i \in \mathcal{S}$ is the full persistent state of the machine after step $i$: application databases, files, and the user interface. An action $a \in \mathcal{A}$ is an operation issued by the agent, including GUI operations, such as click, drag, scroll, typing, or a key combination and CLI commands. Each agent issues actions in its own native grammar; the harness normalizes every action into a shared executable form, so the history of one agent can be replayed for another. At step $i$ the agent observes a screenshot $o_i = \Omega(s_{i-1}) \in \mathcal{O}$, which reveals only part of the state, and its action updates the state as $s_i = \mathcal{T}(s_{i-1}, a_i)$.
We treat $\mathcal{T}$ as deterministic once reproducibility-breaking factors such as automatic updates and notification daemons are disabled; this is what allows a recorded action prefix to be replayed into the same state. A task is a triple $(l, s_0, \mathcal{C})$: a natural-language instruction $l \in \mathbf{L}$, an initial snapshot $s_0$, and a rubric $\mathcal{C} = \{(c_j, w_j)\}_{j=1}^{m}$ of binary criteria with weights $w_j > 0$.

\paragraph{Rollouts and Evaluation.}
\label{sec:rolloutandeval}
At step $i$, a VLM policy $\pi$ samples $a_i \sim \pi(\cdot \mid l, o_i, H_{i-1})$, where $H_{i-1} = (o_1, a_1, \dots, o_{i-1}, a_{i-1})$ represents the interaction history. The episode ends when the agent emits $\texttt{terminate}$ or exhausts its step budget $B$, yielding a rollout $\xi = (o_1, a_1, \dots, o_T, a_T)$ and a final environment state $s_T$.
The rollout records observations because states are only partially observable; since $\mathcal{T}$ is deterministic, any state $s_i$ can be reconstructed by replaying $a_1, \dots, a_i$ from $s_0$. 
An action can mutate invisible state components whose effects persist and influence subsequent operations. A wrong action may thus remain latent for many steps until explicitly revealed by a future observation. Together with this issue, most agent frameworks constrain the context window to the $K$ most recent screenshots ($K = 20$ in our evaluation). Consequently, visual evidence can drop out of the agent's context while its persistent state effects remain active. Each evaluation criterion $c_j(l, \xi) \in \{0, 1\}$ is determined from the rollout, its associated screenshots, and the final state $s_T$. The overall rubric score is defined as $\rho(\xi) = \sum_j w_j c_j / \sum_j w_j$. We define a rollout succeeds ($V(\xi) = 1$) if and only if all criteria are satisfied; otherwise, it fails. We run each evaluation three times: a task counts as solved under Pass@3 if any one of three rollout succeeds, and the Rubric Score is the score of that rollouts.

\paragraph{Root cause, takeover depth, and error horizon.}
In a failed rollout, the root cause step $t_r$ is defined as the earliest action that the agent does not undo, and without which the task remained solvable from $s_{t_r-1}$. Its associated error type $\kappa$ follows the taxonomy in Appendix~\ref{sec:taxonomy}. All identified errors arise from the agent's own decisions. The takeover depth $d$ counts the steps taken after the root cause. An erroneous state at depth $d$ is the state $s_{t_r + d}$ reached by replaying the prefix $\xi_{\le t_r + d}$ of a failed rollout, after unrelated earlier mistakes have been repaired so that the prefix contains exactly one root cause (\S\ref{sec:synthesis}). An agent $\pi'$ takes over with $l$ and $H_{t_r + d}$ as its own history, and we score the full trajectory with $V$ and $\rho$. For each failure we also annotate the error horizon $h_e = t_e - t_r$, where $t_e$ is the first step at which the error becomes explicit in the trajectory; $h_e$ is unobserved if the trajectory ends before $t_e$. We write $c = t_r + d$ for the takeover step and $H_c$ for the interaction history up to step $c$. Execution errors become visible in the next observation ($h_e \ge 1$), whereas planning and perception errors are implicit in the history from $t_r$ and can surface much later.

\section{The \CUAREC{} Benchmark and ReRail}
\label{sec:benchmark}

Building on the formulation above, we introduce \textbf{\CUAREC{}}, a benchmark for error detection and recovery in long-horizon computer-use tasks, and \textbf{\RERAIL{}}, the pipeline that builds it. \CUAREC{} addresses the three gaps in Section~\ref{sec:intro}: its errors come from the agents' own policies rather than injection (coverage), propagate through dependencies across applications and stay hidden for many steps (horizon), and arise in a persistent desktop environment where agents act through both GUI and CLI (interface). \RERAIL{} composes seed tasks into long multi-application workflows, collects failures from multiple agents, and turns them into \DERAILEVAL{}, a human-verified test set of 2,250 erroneous states, and \WATCHDOG{}, a training set of 8,577 verified recovery trajectories. Figure~\ref{fig:overview} summarizes the pipeline.

\subsection{Building the Benchmark with \RERAIL{}}
\label{sec:synthesis}

The key idea of \RERAIL{} is that every task carries a gold lineage, the correct value at every step. An agent's own failure can then be traced against this lineage to locate its root cause, so no errors need to be injected. We outline the main steps below; Appendix~\ref{app:synthesis} gives the details.

\paragraph{Building tasks with dependencies.}
We start from the tasks in MyPCBench~\citep{jang2026mypcbench}. For each task, \RERAIL{} creates an executable intermediate representation that records the application operations, the data they read and write, and the checks on their outputs. Running this graph on a fresh environment snapshot gives the expected intermediate values and final state. To form longer workflows, we connect two tasks when the output of the first can serve as an input to the second. A value that was given in the original instruction must now be found through earlier work, creating a dependency along which a mistake can spread. The resulting pool contains 477 workflows, with medians of three source tasks and 22 graph nodes per workflow.

\paragraph{Checking the generated tasks.}
We execute each workflow to obtain its reference intermediate values and final state, then freeze its verifiers to this gold lineage. We perturb intermediate values and rerun the workflow; a mutation is caught if a downstream verifier fails or if the final outputs or stored data differ from the gold lineage. We retain workflows that catch at least half of the executed mutations and select tasks across different dependency lengths.

\paragraph{Constructing erroneous states and the evaluation suite.}
We split the composed workflows into training and test tasks, with all rollouts, takeover states, and recovery trajectories derived from the same workflow assigned to one split. We run six agents on the workflows: GPT~5.5, Claude~Opus~4.8, Kimi-K3, Qwen3.5-35B-A3B, EvoCUA-32B, and OpenCUA-72B. For each failed rollout, \RERAIL{} compares the recorded execution with the gold lineage to identify the root-cause step $t_r$, the error type, and the first step at which the error becomes evident. To keep each case controlled, experts fix any unrelated mistakes before $t_r$, so that the root cause is the only error in the prefix. \RERAIL{} then creates takeover states by replaying the rollout to depths $d \in \{0,5,10,15,20,25\}$ after the root cause. Each replay must reproduce the state recorded at the same point in the original rollout. For the test suite, human annotators verify the root cause, error type, error horizon, repaired prefix, and replayed state. These annotations support the awareness metric and analyses by error type and horizon (Appendix~\ref{app:human-annotation-protocol}).

\paragraph{Generating the training suite.}
For each erroneous state in the training split, the base model, Qwen3.5-35B-A3B, attempts to complete the task with guidance about the earlier error. Recovery generation begins with a hint that identifies the erroneous step, then adds the conflicting evidence and the affected data if earlier attempts fail. The first continuation accepted by the final verifier becomes the training target; a teacher model is used when the base model produces no verified continuation. Each training example contains the task instruction, the erroneous history, and the verified continuation. The hints are omitted from the training input, matching the unaware takeover setting used for evaluation. Appendix~\ref{app:recovery-training} describes the hint schedule and generation budget.

\subsection{Dataset Composition and Analysis}
\label{sec:dataset}
\label{sec:test}
\label{sec:splits}
\label{sec:stats}
\label{sec:collection}

\paragraph{Evaluation at different takeover depths.}
At each test state, an agent receives the task and the preceding interaction history, then continues without an error warning. The six takeover depths cover histories in which the root-cause screenshot is recent, near the 20-screenshot context limit, or outside that limit. The test set contains 602, 443, 373, 328, 273, and 231 states at depths $d=0,5,10,15,20,25$, respectively. The number of eligible cases decreases with depth; the replay and eligibility rules are described in Appendix~\ref{app:depthinstantiation}.

\paragraph{Training and test suites.}
Table~\ref{tab:dataset-stats} summarizes the two splits. The training suite provides examples of how to finish a task after an error; the test suite provides replayable states and human-checked labels for measuring error awareness and recovery. These labels also let us examine what kinds of errors agents make and how long those errors remain hidden before becoming evident.

\begin{figure}[t]
\centering
\begin{minipage}[b]{0.47\linewidth}\centering
\includegraphics[width=0.86\linewidth,trim=10bp 29.9bp 213bp 5.5bp,clip]{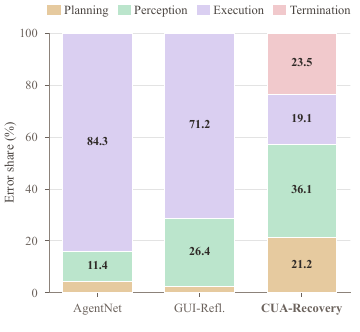}\\[1pt]
{\small (a) Error types}
\end{minipage}\hfill
\begin{minipage}[b]{0.518\linewidth}\centering
\includegraphics[width=0.86\linewidth,trim=195.5bp 29.9bp 10bp 5.5bp,clip]{error_mismatch.pdf}\\[1pt]
{\small (b) Error horizon}
\end{minipage}
\caption{Error types and the delay before an error becomes explicit.}
\label{fig:error_mismatch}
\label{fig:analysis}
\end{figure}

\paragraph{Errors beyond individual actions.}
\label{sec:failures}
Most failures in \CUAREC{} are wrong decisions, not wrong clicks. They span planning, progress and state perception, execution, and termination errors. Examples include choosing records from the wrong year, reading a quantity incorrectly and using it in an order, and creating events in the wrong calendar application (Appendix~\ref{app:examples}). These cases require the agent to check earlier decisions and the data they produced. Planning and perception account for 57.3\% of root causes, termination for 23.5\%, and execution for 19.1\%. For comparison, the available estimates for AgentNet and GUI-Reflection assign 84.3\% and 71.2\% to execution errors (Figure~\ref{fig:error_mismatch}a).

\paragraph{Errors that become clear later.}
Most errors surface only at the end of a rollout. In 602 of the 619 labeled failures, the error becomes explicit within the recorded rollout, and in 361 of them (60.0\%), the evidence first appears at the final step. The median delay from the root cause is 17 steps, and 78.1\% of errors surface at least three steps later. This gives \CUAREC{} a substantial set of failures in which recovery requires revisiting earlier work. Figure~\ref{fig:error_mismatch}b shows this delay distribution; the curves for AgentNet and GUI-Reflection are estimates.

\section{Experiments}
\label{sec:experiments}

We use \CUAREC{} to answer two questions: (1)~can current agents recover from errors they inherit (Section~\ref{sec:main-results}), and (2)~can \WATCHDOG{} teach recovery (Section~\ref{sec:exp-rerail})?

\subsection{Evaluation Setup}
\label{sec:exp-setup}

\paragraph{Agents, metrics, and protocols.}
\label{sec:metrics}
We evaluate GPT~5.5, Claude~Opus~4.8, Kimi-K3, EvoCUA-32B, Qwen3.5-35B-A3B, and OpenCUA-72B. Agents use their respective scaffolds; EvoCUA and OpenCUA use GUI actions only, while the other configurations expose command-line tools where supported (Appendix~\ref{app:baselines}). In self-takeover, each agent resumes states drawn from its own failures. It receives the recorded history without an error warning and can take up to 100 additional steps. Error-Awareness Rate (EAR) measures whether an agent explicitly recognizes a problem in the inherited work. Rubric Score, defined in Section~\ref{sec:rolloutandeval}, measures partial task completion, while Pass@3 reports full completion in at least one of three runs. Reporting them together separates expressed awareness from recovery outcomes.

\paragraph{Clean-start evaluation.}
We also run agents from the initial task state on MyPCBench~\citep{jang2026mypcbench} and Odysseys~\citep{jang2026odysseys}, with 184 and 200 tasks, respectively. Each run is limited to 150 steps and 3,600 seconds. MyPCBench supplies the source tasks used to build our workflows and measures clean-start performance in the source environment. Odysseys contains live-web tasks outside the training environment and measures whether the effect of recovery training extends to a different task setting. 

\subsection{Benchmarking Recovery on \DERAILEVAL{}}
\label{sec:main-results}
\label{sec:findings}
\label{sec:further-analysis}

We let each of the six agents take over states from its own failures at six diverse takeover depths. Table~\ref{tab:recovery-main} reports EAR, Rubric Score, and Pass@3. These results lead to four findings as followed.

\paragraph{Finding 1: Current agents rarely recover completely from their own failures.}
Complete recovery is difficult for all six agents on \DERAILEVAL{}. Across all agents and takeover depths, Pass@3 reaches at most 25.5\%, achieved by GPT~5.5 at $d=20$ (Table~\ref{tab:selftakeover}). Claude~Opus~4.8 reaches at most 7.7\%, while none of the three open-source agents exceeds 6.4\%. These results show that even strong computer-use agents often fail to complete a task once an earlier error has affected the workflow.

\paragraph{Finding 2: Clean-start performance does not reliably predict error awareness.}
Kimi-K3 achieves 66.9\% Pass@3 on MyPCBench from a clean start, but recognizes only 7.8--18.5\% of the inherited errors in its own failures. In contrast, OpenCUA-72B achieves 5.1\% clean-start Pass@3 but recognizes 14.1--26.1\% of its inherited errors (Tables~\ref{tab:clean-start} and~\ref{tab:recovery-main}). 

\paragraph{Finding 3: Recognizing an error rarely leads to recovery.}
Across the self-takeover runs of all six agents, EAR reaches at most 44.0\% and remains below 33\% for all agents except Claude~Opus~4.8 (Table~\ref{tab:recovery-main}). We next group each agent's individual runs by whether it recognizes the inherited error and compute verifier success within each group. Macro-averaged across the six agents, the verifier pass rate is $15.6\%$ for aware runs and $12.2\%$ for unaware runs (Appendix Table~\ref{tab:conditional-awareness}). Most runs still fail after recognizing the error: agents often identify a problem but leave the affected state unchanged or fail to repair all of its consequences. In unaware runs, agents almost always continue from the inherited state without revisiting the earlier work. Awareness is therefore associated with a greater chance of recovery, but recognition alone is rarely sufficient.

\paragraph{Finding 4: Later takeovers remain difficult even when awareness improves.}
\label{sec:depth-analysis}
\label{sec:ooc}
From $d=20$ to $d=25$, the root-cause screenshot moves beyond the 20-screenshot history. Over this interval, GPT~5.5's EAR rises from 16.2\% to 21.0\%, while its Pass@3 falls from 25.5\% to 18.9\%. Claude's EAR also rises, from 18.8\% to 27.3\%, while Pass@3 falls from 7.7\% to 0.0\%. For Kimi-K3, Pass@3 falls from 23.1\% to 9.1\%. This pattern suggests a trade-off at later takeover points. As an error propagates through the workflow, its effects become more visible in the current state, making the problem easier to notice. Once the root-cause screenshot leaves the context window, however, the agent must infer how the problem began from its downstream effects, making the error harder to diagnose and repair. This helps explain why EAR can increase even as Pass@3 falls.

\newcommand{\recUp}[1]{\textbf{#1}}
\newcommand{\recDn}[1]{#1}
\definecolor{recbg}{RGB}{234,244,251}
\newcommand{\recRR}{\hspace{0.5em}$\hookrightarrow$\,\RERAIL{}}

\begin{table}[h]
\caption{Recovery on \DERAILEVAL{} by takeover depth $d$ (steps after the root cause). Each agent takes over states from its \emph{own} failures without an error warning. \textbf{$\hookrightarrow$\,\RERAIL{}: \RERAILFULL{}} takes over the same states as the agent above. Opus 4.8 = Claude Opus 4.8; Qwen3.5-35B = Qwen3.5-35B-A3B, the base model of \RERAIL{}. \textbf{Bold}: higher than the corresponding source agent.}
\label{tab:recovery-main}
\centering
\small
\setlength{\tabcolsep}{1.55pt}\renewcommand{\arraystretch}{1.18}
\begin{tabular}{@{}l cccccccccccc@{}}
\toprule
& \multicolumn{6}{c}{Error Awareness (\%)} & \multicolumn{6}{c}{Rubric Score / Pass@3 (\%)} \\
\cmidrule(lr){2-7}\cmidrule(l){8-13}
Agent \hfill $d{=}$ & 0 & 5 & 10 & 15 & 20 & 25 & 0 & 5 & 10 & 15 & 20 & 25 \\
\midrule
\textit{API-based} &&&&&&&&&&&& \\
GPT 5.5 & 20.1 & 26.7 & 32.3 & 28.6 & 16.2 & 21.0 & 65.3/20.1 & 67.7/24.3 & 61.2/20.9 & 60.8/20.1 & 68.5/25.5 & 54.9/18.9 \\
Opus 4.8 & 18.9 & 44.0 & 39.1 & 22.7 & 18.8 & 27.3 & 35.3/0.0 & 24.0/0.0 & 29.5/4.5 & 38.5/5.3 & 48.1/7.7 & 36.4/0.0 \\
Kimi-K3 & 13.9 & 7.8 & 10.0 & 11.1 & 18.5 & 15.0 & 39.2/11.6 & 41.7/8.6 & 36.4/7.4 & 33.1/11.8 & 41.5/23.1 & 36.9/9.1 \\
\rowcolor{recbg}\recRR & \recUp{25.0} & \recUp{15.7} & \recUp{17.5} & \recUp{22.9} & \recDn{14.8} & 15.0 & \recDn{36.4}/\recUp{16.6} & \recDn{34.8}/\recUp{12.6} & \recDn{31.7}/\recUp{10.1} & \recDn{32.4}/\recDn{11.6} & \recDn{36.8}/\recDn{15.2} & \recDn{30.0}/\recDn{8.8} \\
\midrule
\textit{Open-source} &&&&&&&&&&&& \\
EvoCUA-32B & 24.7 & 28.3 & 30.3 & 30.4 & 32.7 & 21.8 & 25.6/6.4 & 27.1/6.1 & 23.8/4.4 & 19.3/3.7 & 24.5/4.6 & 16.1/2.3 \\
\rowcolor{recbg}\recRR & \recUp{45.3} & \recUp{43.9} & \recUp{42.4} & \recUp{49.5} & \recUp{44.6} & \recUp{37.7} & \recUp{29.5}/\recUp{12.7} & \recUp{31.6}/\recUp{8.8} & \recUp{26.1}/\recUp{6.9} & \recUp{24.3}/\recUp{5.4} & \recUp{28.1}/\recDn{3.1} & \recUp{18.5}/\recUp{5.4} \\
Qwen3.5-35B & 25.8 & 21.4 & 23.9 & 18.5 & 14.3 & 15.6 & 8.4/1.2 & 6.5/0.0 & 4.2/0.0 & 2.1/0.0 & 4.6/0.0 & 3.9/0.0 \\
\rowcolor{recbg}\recRR & \recUp{41.5} & \recUp{37.0} & \recUp{38.6} & \recUp{34.2} & \recUp{30.8} & \recUp{29.0} & \recUp{14.3}/\recUp{4.7} & \recUp{18.5}/\recUp{3.5} & \recUp{17.4}/\recUp{2.7} & \recUp{18.1}/\recUp{1.8} & \recUp{13.8}/0.0 & \recUp{14.3}/0.0 \\
OpenCUA-72B & 26.1 & 20.4 & 22.4 & 14.1 & 19.0 & 19.4 & 6.0/0.9 & 4.8/0.0 & 3.4/0.0 & 4.7/0.0 & 3.6/0.0 & 5.7/0.0 \\
\bottomrule
\end{tabular}
\end{table}

\subsection{Training Recovery with \WATCHDOG{} and Generalization}
\label{sec:exp-rerail}
\label{sec:cross}
\label{sec:generalization}

The previous results show that agents have separate difficulties in noticing an inherited error and repairing its downstream effects. We next examine whether the verified trajectories in the training suite can support recovery-oriented fine-tuning using a standard supervised objective.

\paragraph{Training setup.}
We fine-tune Qwen3.5-35B-A3B for one epoch on the 8,577 verified recovery trajectories in \WATCHDOG{}, obtaining \textbf{\RERAILFULL{}}. Given a task instruction $l$, an erroneous history $H_c$, and its verified recovery $r^\star$, we minimize $$\mathcal{L}(\theta)=-\mathbb{E}_{(l,H_c,r^\star)\sim\mathcal{Q}}[\log \pi_\theta(r^\star \mid l,H_c)]$$, where $\mathcal{Q}$ is the training set. The hints used to generate $r^\star$ are not included in $H_c$, so the model receives the same unaware input used during evaluation. 

\paragraph{Finding 5: Recovery learned from \WATCHDOG{} transfers across agents.}
We first evaluate the fine-tuned model on states taken from Qwen3.5-35B-A3B's own failures, allowing the two models to start from the same erroneous states. Fine-tuning raises EAR by 13.4--16.5 percentage points and Rubric Score by 5.9--16.0 points across takeover depths (Table~\ref{tab:recovery-main}). At $d=0$, Pass@3 increases from 1.2\% to 4.7\%. EAR and Rubric Score also improve at $d\ge20$, although neither model solves any task at these depths. The training data therefore improves both error recognition and partial recovery, while complete recovery from later takeover points remains difficult. We also evaluate the fine-tuned model on failures produced by Kimi-K3 and EvoCUA-32B. At $d=0$, \RERAILFULL{} reaches 16.6\% Pass@3 on Kimi-K3's failure states, compared with 11.6\% for Kimi-K3, and 12.7\% on EvoCUA-32B's states, compared with 6.4\% for EvoCUA-32B (Figure~\ref{fig:cross}). Its advantage over Kimi-K3 holds through $d=10$, and its advantage over EvoCUA-32B holds through $d=15$. At later depths, these gains narrow or disappear. The results show that the learned recovery behavior transfers across failure sources, with the largest gains at earlier takeover points.

\begin{table}[h]
\caption{Completion performance on MyPCBench and Odysseys. Bold and underline indicate the best and second-best results within API-based and open-source agents.}
\label{tab:clean-start}
\centering
\small
\resizebox{0.95\linewidth}{!}{
\begin{tabular}{@{}lrrrr@{}}
\toprule
 & \multicolumn{2}{c}{MyPCBench} & \multicolumn{2}{c}{Odysseys} \\
\cmidrule(lr){2-3}\cmidrule(l){4-5}
Agent & Pass@3 (\%) & Rubric Score (\%) & Pass@3 (\%) & Rubric Score (\%) \\
\midrule
\multicolumn{5}{@{}l}{\emph{API-based}} \\
GPT~5.5 & \textbf{85.0} & \textbf{92.2} & \textbf{82.1} & 94.2 \\
Claude Opus~4.8 & \underline{68.9} & \underline{84.6} & \underline{81.7} & \textbf{95.6} \\
Kimi-K3 & 66.9 & 83.6 & 77.9 & \underline{94.3} \\
\midrule
\multicolumn{5}{@{}l}{\emph{Open-source}} \\
EvoCUA-32B & \underline{18.1} & \underline{41.0} & \underline{32.2} & \underline{73.4} \\
Qwen3.5-35B-A3B & 9.2 & 33.4 & 24.1 & 66.5 \\
OpenCUA-72B & 5.1 & 20.5 & 18.9 & 52.1 \\
\rowcolor{oursrow}\textbf{\RERAILFULL{}} &
\eargain{\textbf{24.5}}{15.3} &
\eargain{\textbf{53.2}}{19.8} &
\eargain{\textbf{34.9}}{10.8} &
\eargain{\textbf{78.2}}{11.7} \\
\bottomrule
\end{tabular}
}
\end{table}

\paragraph{Finding 6: Recovery training also improves task completion from a clean start.}
Training on \WATCHDOG{} raises Pass@3 from 9.2\% to 24.5\% and Rubric Score from 33.4\% to 53.2\% on in-domain MyPCBench. On Odysseys, whose tasks and applications are outside the training environment, the scores rise from 24.1\% to 34.9\% and from 66.5\% to 78.2\%, respectively, showing gains beyond the source environment (Table~\ref{tab:clean-start}).

\paragraph{Ablation study.}
\label{sec:ablation}
To test whether the gains come from recovery supervision or simply from more in-domain training, we fine-tune the same base model on a token-matched control of successful trajectories from the same workflows. On Qwen3.5-35B-A3B's failures, averaged over depths, EAR rises from 19.9\% to 23.6\% with successful trajectories but to 35.2\% with recovery trajectories, and clean-start Pass@3 on MyPCBench reaches 13.8\% versus 24.5\% (Appendix~\ref{app:ablation}). Recovery trajectories thus bring much larger gains than successful trajectories from the same workflows, showing that the improvement comes from learning to recover rather than from more in-domain data.

\section{Conclusion}
We find that current computer-use agents rarely recover from their own errors in long-horizon workflows, and strong performance does not imply reliable recovery. Existing data and benchmarks miss this setting in three ways: errors are injected or low-level rather than agent-induced, trajectories are too short for errors to stay hidden, and evaluation covers only GUI actions with screenshot-based scoring. We present \CUAREC{}, a benchmark built with \RERAIL{} from agents' own failures in long multi-application workflows involving both GUI and CLI, with a human-verified test set and verified recovery trajectories for training. Fine-tuning on \WATCHDOG{} improves recovery from inherited errors and clean-start performance, including in an unseen environment, suggesting that learning from real failures is a practical path toward reliable computer-use agents.

\bibliography{iclr2027_conference}
\bibliographystyle{iclr2027_conference}

\appendix

\addtocontents{toc}{\protect\setcounter{tocdepth}{3}}
\clearpage
\tableofcontents
\clearpage

\section{Cost}
\label{app:cost}

Tables~\ref{tab:cost}, and~\ref{tab:cost-gpu} itemize the API, and GPU cost of the paper. Judging one group with GPT~5.6-terra , either one run of one agent's $184$ clean-start rollouts or one agent--horizon cell of takeover episodes, costs about \$120. Prefix replay executes recorded actions without model calls, so the API cost of a takeover comes only from the agent acting after the takeover point; we estimate it from the per-task cost of the same agent's clean-start run. Ten annotators labeled the failure corpus (Appendix~\ref{app:annotation-staffing}) at \$100 per annotator per labeling group. Locally served agents and \RERAILFULL{} incur no API cost; each serves one episode at a time on eight AMD MI308X GPUs (Appendix~\ref{app:agents}), so running the $32$ concurrent sandboxes against local models occupies up to $256$ GPUs. 

\begin{table}[H]
\caption{API cost. A judge group is one run of one agent's clean-start rollouts ($3$ runs per agent) or one
agent--horizon cell of takeover episodes. Red values are estimates.}
\label{tab:cost}
\centering
\small
\setlength{\tabcolsep}{5pt}
\resizebox{\linewidth}{!}{%
\begin{tabular}{@{}llrrr@{}}
\toprule
Item & Unit & Unit cost (USD) & Count & Total (USD) \\
\midrule
\multicolumn{5}{@{}l}{\emph{Clean-start collection}} \\
API agent rollouts (Claude~Opus~4.8, GPT~5.5; 3 runs) & run & 260 & 6 & 1{,}560 \\
API agent rollouts (Kimi-K3; 3 runs) & run & 252 & 3 & 756 \\
Judge (GPT~5.6 \texttt{terra}) & group & 120 & 18 & 2{,}160 \\
\midrule
\multicolumn{5}{@{}l}{\emph{Self-takeover on \DERAILEVAL{} (Table~\ref{tab:selftakeover}; 6 agents $\times$ 6 horizons, unaware)}} \\
Claude~Opus~4.8 takeover execution & episode & 1.41 & 139 & 196 \\
GPT~5.5 takeover execution & episode & 1.41 & 69 & 97 \\
Kimi-K3 takeover execution & episode & 1.37 & 175 & 240 \\
Judge (GPT~5.6 \texttt{terra}) & group & 120 & 36 & 4{,}320 \\
\midrule
\multicolumn{5}{@{}l}{\emph{Cross-policy takeover by \RERAILFULL{} (Tables~\ref{tab:recovery-main} and~\ref{tab:training-awareness}; 3 sources $\times$ 6 horizons)}} \\
Judge (GPT~5.6 \texttt{terra}) & group & 120 & 18 & 2{,}160 \\
\midrule
\multicolumn{5}{@{}l}{\emph{Data construction}} \\
Task IR extraction (GPT~5.6) & call & 0.46 & 483 & 224 \\
Instruction realization (GPT~5.6) & call & 0.034 & 1{,}780 & 60 \\
Recovery generation for \WATCHDOG{}: teacher fallback & attempt & 1.41 & 3{,}350 & 4{,}720 \\
\midrule
API total & & & & 16{,}493 \\
\bottomrule
\end{tabular}%
}
\end{table}

\begin{table}[H]
\caption{GPU and sandbox compute. Each locally served model runs one episode
at a time on a server of eight AMD MI308X GPUs, so GPU-hours are eight times
episode wall-clock time; API agents use sandbox time only. Clean-start times
are measured; takeover times use each agent's measured per-step latency over
58 post-takeover steps.}
\label{tab:cost-gpu}
\centering
\small
\setlength{\tabcolsep}{5pt}
\begin{tabular}{@{}lrrrr@{}}
\toprule
Agent & Episodes & Min./episode & GPU-hours & Sandbox-hours \\
\midrule
\multicolumn{5}{@{}l}{\emph{Clean-start collection}} \\
Claude~Opus~4.8 & 552 & 37.6 & -- & 345 \\
GPT~5.5 & 552 & 9.2 & -- & 84 \\
Kimi-K3 & 552 & 30.5 & -- & 282 \\
EvoCUA-32B & 552 & 25.1 & 1{,}872 & 234 \\
OpenCUA-72B & 552 & 7.6 & 558 & 69 \\
Qwen3.5-35B-A3B & 552 & 15.9 & 1{,}170 & 147 \\
\midrule
\multicolumn{5}{@{}l}{\emph{Self-takeover (Table~\ref{tab:selftakeover})}} \\
Claude~Opus~4.8 & 139 & 69.5 & -- & 161 \\
GPT~5.5 & 69 & 10.2 & -- & 12 \\
Kimi-K3 & 175 & 49.4 & -- & 144 \\
EvoCUA-32B & 613 & 15.9 & 1{,}296 & 162 \\
OpenCUA-72B & 683 & 7.5 & 687 & 86 \\
Qwen3.5-35B-A3B & 571 & 10.1 & 765 & 96 \\
\midrule
\multicolumn{5}{@{}l}{\emph{Cross-policy takeover by \RERAILFULL{} (Tables~\ref{tab:recovery-main} and~\ref{tab:training-awareness})}} \\
\RERAILFULL{} & 1{,}359 & 10.1 & 1{,}822 & 228 \\
\midrule
\multicolumn{5}{@{}l}{\emph{Fine-tuning (Table~\ref{tab:train-details})}} \\
\RERAILFULL{} training, 64 GPUs $\times$ 18 hours & -- & -- & 1{,}152 & -- \\
\midrule
Total & & & 9{,}322 & 2{,}050 \\
\bottomrule
\end{tabular}
\end{table}

\section{Details of the \RERAIL{} Data-Synthesis Framework}
\label{app:synthesis}

This appendix gives the \RERAIL{} framework of Section~\ref{sec:synthesis} in
full. Although the current MyPCBench task set offers realistic workflows,
most of the tasks are relatively short, requiring only 20 to 45 steps for a
capable agent to complete, and the dependency structures that allow errors
to remain hidden typically span no more than two nodes. Studying hidden
errors at scale requires tasks whose dependency chains are long and whose
ground truth is executable, so that every intermediate value, every written
cell, and the final state are known before any agent runs, properties that
current benchmarks lack. We therefore synthesize tasks by composing verified
source tasks and accept a composition only after it has been executed and
its verifiers have been shown to reject wrong runs. This appendix describes
the pipeline as implemented: the extraction of an executable task
representation from each source task (\S\ref{sec:synthesis-ir}), the
composition search with an executable gate (\S\ref{sec:synthesis-search}),
verifier freezing and the mutation test that measures verifier coverage
(\S\ref{sec:synthesis-verifier}), horizon-controlled selection with world and
instruction variants (\S\ref{sec:synthesis-select}), and the scale and horizon
of the released set (\S\ref{sec:synthesis-scale}).

\subsection{From Workflows to Training and Test Examples}
\label{app:rerail-procedure}

Algorithm~\ref{alg:rerail} summarizes how \RERAIL{} uses validated workflows to construct the two splits.

\begin{algorithm}[h]
\caption{\RERAIL{}: collecting failures and building recovery data}
\label{alg:rerail}
\footnotesize
\begin{algorithmic}[1]
\Require validated task pool $\mathcal{M}$ with reference executions $G(M)$ and frozen verifiers; source policies $\Pi$; base model $\pi_0$; teacher $\pi_{\mathrm{T}}$; depths $\mathcal{D}=\{0,5,10,15,20,25\}$
\Ensure training recovery set $\mathcal{Q}$ and test-state set $\mathcal{E}$
\State Split $\mathcal{M}$ into training and test tasks
\State $\mathcal{Q},\mathcal{E}\gets\emptyset$
\For{task $M\in\mathcal{M}$, instruction $l$ for $M$, source policy $\pi\in\Pi$}
\State $\xi\gets\textsc{Rollout}(\pi,l,M)$
\State \textbf{if} the final verifier accepts $\xi$ \textbf{then continue}
\State $(t_r,\kappa,t_e)\gets\textsc{AnalyzeFailure}(\xi,G(M))$
\State \textbf{if} the root cause is unresolved or type confidence is below $0.7$ \textbf{then continue}
\State $\xi\gets\textsc{RepairEarlierMistakes}(\xi,t_r)$
\State \textbf{if} prefix repair fails \textbf{then continue}
\For{$d\in\mathcal{D}$ with $c=t_r+d\le|\xi|$}
\State \textbf{if} replay does not reproduce the recorded state \textbf{then continue}
\If{$M$ is a test task}
\State Apply the test eligibility rules of Appendix~\ref{app:depthinstantiation}
\State Obtain human checks of the labels and replayed state
\State Add accepted state, history, and labels to $\mathcal{E}$
\Else
\State $r^\star\gets$ first verified continuation under the hint schedule of Appendix~\ref{app:recovery-training}
\State \textbf{if} $r^\star\neq\bot$ \textbf{then} add $(l,H_c,r^\star)$ to $\mathcal{Q}$, without the hint
\EndIf
\EndFor
\EndFor
\State \Return $\textsc{Dedup}(\mathcal{Q}),\mathcal{E}$
\end{algorithmic}
\end{algorithm}

\subsection{From Source Tasks to Executable Task IR}
\label{sec:synthesis-ir}

\paragraph{Task Intermediate Representation.}
A task is represented as a grounded DAG $M=(V,E)$ over one frozen environment
snapshot. Every node $v\in V$ records the application it acts on, and its operation
is selected from the following action set: retrieve, resolve, filter, compare, aggregate, verify,
decide, create, modify, delete, communicate, confirm. It also defines the table cells it reads,
the values it produces along with an executable derivation, such as a read-only SQL query,
an expression over upstream values, a file read, or a row or file creation,
the cells it writes, its reversibility class, and a node verifier.
An edge transfers a value or state from an output port to an input port; if no edge connects to a specific input port, that port holds the literal or constant specified in the instruction. The graph carries a final-state verifier with an explicit set of equivalent final states. The snapshot consists of 17 application
databases with 201 tables, a schema graph that captures both foreign-key and inferred
relationships, and an inventory of the 48 documents in the persona's home directory.
The file inventory is exposed as a single virtual table, allowing files and database
rows to be treated as the same kind of fact.

\paragraph{Extraction and programmatic validation.}
A language model extracts the Task IR of each of the 184 MyPCBench source tasks
from the instruction, the rubric, the persona, and the schema. The result is validated programmatically rather than by an LLM judge. Static checks verify that the DAG is valid, that every cell is grounded in the schema graph, and that every named entity exists in the snapshot. Each input literal must either appear in the instruction or refer to the persona's own identity, ensuring that facts the agent is expected to retrieve are not provided directly as constants.

The gold IR is then executed on a fresh copy of the snapshot. The interpreter evaluates each derivation, performs the declared writes, checks the cells actually read by each SQL query against the declared reads, and aborts if it encounters an undeclared write. It also records the gold lineage $G(M)$, including every produced value, every written cell, and the state digest after each node.

Finally, the gold values are compared with the ground truth in the source rubric, such as amounts, dates, and names. Relative-date variables are expanded using the snapshot's reference time. When a rubric contains no comparable literals, it is evaluated based on the overlap between its wording and the gold values. Any task that fails validation is re-extracted with the program's findings appended to the prompt, for up to four rounds.

We finally accepted 135 of 184 tasks (73.4\%) from the original MyPCBench, based on the validation situation:
53 by literal agreement with the rubric and 82 by wording. Of the 49 remaining tasks, 34 cannot be executed because the model's derivations or writes are incompatible with the actual schema, 8 have rubrics that contain no comparable values, 5 have final verifiers that reject their own gold runs, and 2 contradict their rubrics.
The accepted IRs have a median of 6 nodes. Every accepted IR ships with its
gold lineage, which is the single source of truth for everything that follows.

\subsection{Composition Search with an Executable Gate}
\label{sec:synthesis-search}

\paragraph{Typed compatibility with cell evidence.}
Composition connects an output port \(p_o\) of one accepted IR to a literal-valued input port \(p_i\) of another. The value originally provided as a constant in the second task's instruction must then be obtained by executing the first IR.

Let \(\tau(p_o,p_i)\in\{0,0.4,1\}\) denote the type-compatibility score between the two ports. The score is \(1\) when their entity-reference types match, \(0.4\) when they share a generic scalar type such as Integer or Text, and \(0\) otherwise. For each port $p$, let $\mathrm{col}(p)$ denote the schema column associated with that port. We first identify $\mathrm{col}(p)$ from the IR's declared reads. If the declared reads do not identify a unique column, we recover it from the SQL derivation. For an output port, $\mathrm{col}(p)$ is the only column selected by the scalar query. For an input port, it is the column compared with the query parameter corresponding to $p$. If no unique column can be identified, the port is not eligible for composition.

Let $\mathrm{rel}(p_o,p_i)$ indicate that the tables containing $\mathrm{col}(p_o)$ and $\mathrm{col}(p_i)$ are connected in the schema graph by a key path of at most two hops.
The compatibility score is
\begin{equation}
 \sigma(p_o,p_i)=
 \begin{cases}
   1, & \mathrm{col}(p_o)=\mathrm{col}(p_i),\\
   \max\bigl(\tau(p_o,p_i),\,0.8\bigr), & \tau(p_o,p_i)>0\wedge \mathrm{rel}(p_o,p_i),\\
   \tau(p_o,p_i), & \text{otherwise}.
 \end{cases}
 \label{eq:compat-score}
\end{equation}
We call a port pair admissible, written $p_o\leadsto p_i$, if and only if $\sigma(p_o,p_i)\ge 0.5$.
A pair of ports with matching generic scalar types is therefore admissible only if the ports are grounded in the same schema column or their columns are connected by a key path. This additional constraint prevents matches based on type alone. The ports must also have matching cardinalities, and literals representing the persona's identity cannot be used as graft targets. Across the 135 accepted IRs, 1{,}053 output ports and 109 candidate input ports produce 143 admissible edges: 51 connect ports grounded in the same column, and 92 connect type-compatible ports supported by a key path.

\paragraph{Operators and search.}
For each seed IR, we consider beam search for compositions with any IR in the accepted pool. At each step, the search can pass an output of task $A$ to a literal input of task $B$, add another instance of the same operation pattern, attach an additional producer to a comparison or aggregation node, reuse a value several nodes after it is produced, or use a boolean produced by a decide node in $A$ to guard the first state-changing node in $B$. The resulting graph is re-evaluated after each step.

To avoid naming conflicts, each node identifier is prefixed with its parent task. The provenance record for each composition stores the operator, the port binding, and the source tasks.

Before execution, each candidate must pass a set of static checks. A graph is rejected if more than 25\% of it consists of independent components, which usually indicates that two tasks have been placed side by side rather than meaningfully combined. It is also rejected if a value passed from one task to another has no effect on later computation, if fewer than two nodes fall within the target horizon range (\S\ref{sec:synthesis-select}), or if the graph contains fewer than 10 or more than 24 nodes. The lower bound excludes tasks that are too short to meaningfully test long-horizon reasoning (\S\ref{sec:synthesis-scale}). The upper bound keeps tasks within reach of current agents.

Candidates that survive these checks are executed on a fresh copy of the snapshot. The execution must complete successfully, and the final verifier must accept the resulting gold state. The beam retains 16 states, allows at most three grafts, and uses no more than 48 gold executions for each seed in each round.

A full copy of every database is used for each world, so recomputing all database digests after every step would be expensive. Instead, the interpreter caches the digest of each database and recomputes it only when the corresponding connection reports a change. This reduces the time required for one gold execution from 8.9\,s to 0.1\,s, making executable validation practical.

\paragraph{Composed rubrics.}
For a workflow composed from source tasks $T_1,\ldots,T_k$, we form its rubric by combining the rubric items of the source tasks. When composition replaces a source-task input with an upstream output, we update the corresponding expected values and verification descriptions using the composed gold lineage. Each retained item keeps its source-task weight. The resulting rubric is used both to construct the frozen programmatic verifier and, item by item, by the GPT~5.6-Terra rubric judge.

\subsection{Verifier Freezing and the Mutation Test}
\label{sec:synthesis-verifier}

\paragraph{Freezing verifiers to the gold.}
The verifiers produced during extraction mainly check internal consistency rather than correctness. Among the 135 accepted final verifiers, 122 are derived predicates, and 76 refer only to values computed during the run. Such a verifier may therefore accept a run whose outputs are mutually consistent but incorrect. The problem becomes more serious after composition, because the verifier inherited from the receiving task may still contain literals from the original task.

To prevent this, we anchor each verifier to the gold lineage before using it. Let $\mathrm{Sink}(M)$ denote the sink nodes of $M$, whose outputs are not consumed by any other node, and let $g_{u,x}$ denote the gold value of output $x$ from node $u$. Suppose that a verifier predicate $\varphi$ refers to run-time values $\mathrm{val}[u][x]$ and to SQL parameters derived from those values. We freeze the predicate by replacing every reference to a non-sink value with its gold counterpart:
\begin{equation}
\hat{\varphi}(\xi;G(M))
=
\varphi\!\left(
\xi;\,
\mathrm{val}[u][x]\leftarrow g_{u,x}
\text{ for every referenced non-sink value }(u,x)
\right).
\label{eq:freeze}
\end{equation}

The sink values remain unchanged and are taken from the run being evaluated, while the values used to derive the expected result are fixed to the gold lineage. The resulting predicate therefore checks the run's final outputs against a gold-grounded expectation. For a node-level verifier, the node being checked is treated as the sink, so its own outputs remain unfrozen while the values on which the check depends are fixed to gold.

There is one additional minor step in the composition process: before fixing the validators passed down from the receiving task, we must first replace any literals that coincide with the reference answers at the insertion point or subsequent stages with parameters. Any candidate final conditions that still rely on those discarded literals are simply dropped. Next, we concatenate the final validators from the two source tasks, execute the composed IR, and obtain its reference answer lineage. Using this lineage, we then fix the new parameters according to Equation~\ref{eq:freeze}. Finally, we run the fixed IR again using a fresh snapshot; only those that pass this validation are truly included in the benchmark dataset.

\paragraph{Mutation test.}
The mutation test evaluates how effectively verifiers detect incorrect behavior. For each target value, the framework selects an appropriate mutation from a predefined library based on its type. Common mutations include swapping table or entity names, shifting dates by a day or week, scaling amounts, omitting fees, altering list elements, toggling booleans, or using stale text. The IR is then re-executed with the mutated value overriding the original. A mutation $\mu$ at node $v$ is considered caught if any downstream verifier fails, the frozen final verifier fails, or the run's outcome deviates from the gold standard. This final check mimics a human judge comparing the model's output against the grading rubric. The rejection rate of a task is
\begin{equation}
 R(M)=\frac{\bigl|\{\mu\in\mathrm{Mut}(M):\ \mathrm{caught}(\mu)\}\bigr|}{|\mathrm{Mut}(M)|}.
 \label{eq:rejection-rate}
\end{equation}
We aggregate $R(M)$ strictly over mutations that executed, retaining a task in the pool only if $R(M)\ge 0.5$. The first node verifier that fails reveals the mutation's dynamic latent horizon, the topological distance from node $v$ to that failing verifier. Mutations caught exclusively by the final verifier or gold divergence are labeled verifier-only, while those that escape detection entirely are silent. In one generation run with 17{,}857 mutation executions across 417 candidates, we observed a median $R$ of 0.64 (mean 0.67), with 327 candidates passing the threshold. Specifically, downstream node verifiers caught 4{,}711 mutations, final verifiers caught 823, sink or write divergences caught 6{,}394, and 5{,}929 remained silent. Omitting the freezing step and the divergence check drops the median $R$ to 0, leaving only 3 out of 417 candidates alive. 

\subsection{Horizon Buckets, Selection, and Variants}
\label{sec:synthesis-select}

\paragraph{Static Latent Horizon.}
To quantify how deeply an error can hide before being exposed, we compute the static latent horizon $\lambda(v)$ on the IR prior to execution. Conceptually, when a node $v$ produces an incorrect value, it contaminates the data facts it touches. The horizon $\lambda(v)$ measures the shortest path along the dependency graph from $v$ to the first downstream node $u$ that consumes any of these contaminated facts:
\begin{equation}
 \lambda(v)=\min\{\,\mathrm{dist}(v,u):\ u\succ v,\ \mathrm{facts}(u)\cap\mathrm{cont}(v)\neq\emptyset\,\}.
 \label{eq:static-horizon}
\end{equation}
We set $\lambda(v)=\infty$ if the error is only caught by a final verifier, which is a verifier-only error, and leave it undefined if no downstream component detects it, which is a silent error. We group nodes into four categories: $1$, $2$--$3$, $\ge 4$, and verifier-only, and assign each task to the category that most frequently occurs among its nodes. Bucket $1$ represents immediate error propagation where mistakes are caught at the very next step, whereas the higher buckets capture harder cases where errors persist over multiple actions or until the end of execution.

\paragraph{Selection.}
To build a balanced benchmark, we first select base tasks to populate the four horizon buckets according to fixed target shares $\boldsymbol{\alpha}=(0.20, 0.35, 0.30, 0.15)$, yielding per-bucket quotas $q_b = N \, \alpha_b$. Candidates are selected in order of a composite score that rewards a higher proportion of target buckets and cross-application edges, while penalizing dependency depths outside the $[5,9]$ range. Selection is subject to two constraints: source tasks are restricted to the control bucket, and unfilled quotas are recorded as shortfalls rather than backfilled from other buckets.

\paragraph{Axis 1: World variants.}
Once base tasks are selected, we introduce environmental hazards as the first dimension of variation. These hazards include identity aliases, stale documents, title-similar distractor events, and broken preconditions. To prevent artificial over-injection, we first query each hazard's baseline occurrence rate in the snapshot. We inject a hazard only if its baseline rate is non-zero, capping the post-injection rate at twice the baseline. When a node's instruction constant or output matches the hazard selector, the environment seeder applies the hazard as a persona patch. In our snapshot, only distractor events met these conditions (baseline count: 19), whereas identity aliases and stale documents did not qualify for injection.

\paragraph{Axis 2: Instruction variants.}
Instruction realization forms the second dimension of variation. For each selected task, the model generates candidate instructions from the graph summary and public context without revealing restricted ``hidden values'', which are gold values excluded from initial inputs and prior seeds. An instruction is retained only if it contains all required input literals without leaking hidden values. Distinct valid instructions yield independent rollout starting points while retaining the task's core gold target and verifiers.

\subsection{Scale and Horizon of the Synthesized Set}
\label{sec:synthesis-scale}
\begin{table}[H]
\centering
\small
\caption{The synthesis funnel on one generation run over the 135 accepted
source IRs.}
\label{tab:synthesis-funnel}
\begin{tabular}{lr}
\toprule
Stage& Count \\
\midrule
Composed candidates reaching the executable gate & 364 \\
Rejected by a static gate & 82 \\
Candidates with a gold run (composed + source) & 417 \\
Mutation executions & 17{,}857 \\
Candidates with $R(M)\ge 0.5$ & 327 \\
Selected tasks (composed / source) & 277 / 10 \\
\quad by bucket: $2$--$3$ / $1$ / verifier-only / $\ge 4$ & 160 / 76 / 40 / 11 \\
\midrule
Released set & \\
\midrule
Distinct composed tasks & 477 \\
\quad by bucket: $2$--$3$ / $1$ / verifier-only / $\ge 4$ & 296 / 120 / 48 / 13 \\
\quad by number of grafts: 1 / 2 / 3 & 124 / 180 / 173 \\
\bottomrule
\end{tabular}
\end{table}

Across three synthesis runs with increasing beam widths, our pipeline ultimately produced a released set of 477 distinct composed tasks (Table~\ref{tab:synthesis-funnel} gives the detailed synthesis funnel). These tasks exhibit substantial structural complexity, combining a median of 3 source tasks into a 22-node dependency graph with dense cross-application interactions. Among synthesis operators, conditionalization is the most frequent pattern. Furthermore, over 90\% of generated instruction drafts passed our leakage filters, providing most tasks with two validated, independent rollout entry points.

This node density underpins the long-horizon nature of the synthesized set. Empirical rollouts reveal that competent execution consumes roughly 4 to 7 GUI steps per node (25--42 steps for a 6-node source task). Consequently, a median 22-node synthesized task translates to 85--150 steps of efficient tool use. 

\subsection{Statistics of CUA-Recovery}

\begin{table}[h]
\centering
\caption{Statistics of \CUAREC{}. Training examples are verified recovery trajectories; test examples are erroneous takeover states.}
\label{tab:dataset-stats}
\footnotesize
\setlength{\tabcolsep}{4pt}
\begin{tabular}{@{}lrr@{}}
\toprule
 & Train & Test \\
\midrule
Tasks & 377 & 100 \\
Examples & 8{,}577 & 2{,}250 \\
Mean steps per trajectory & 60.0 & 59.7 \\
Multi-application trajectories (\%) & 67.5 & 68.4 \\
Label checks & Automatic & Human \\
\bottomrule
\end{tabular}
\end{table}

\section{More Details of CUA-Recovery}
\label{app:cua_recovery_test}

\subsection{Test-Case Construction}
\label{app:construction}
\label{app:errstate}

Test tasks run in the MyPCBench environment~\citep{jang2026mypcbench}, a persistent personal workspace with up to 17 applications.
Each task has an instruction, a setup routine that restores the initial state, and the frozen verifier of Section~\ref{sec:synthesis}.
We turn each failed rollout into test cases in four steps.

\paragraph{Normalized trajectories.}
\label{app:normalization}
The six agents use different action formats.
We store every step in one shared executable format and convert it back to the native format of the agent that takes over.

\paragraph{Prefix repair.}
\label{app:prefixrepair}
Each test case has exactly one root cause.
\RERAIL{} repairs every unrelated mistake before the root-cause step, and annotators confirm that the repaired prefix has no other error.

\paragraph{Replay verification.}
\label{app:replay}
We restore the task snapshot and replay the repaired prefix, the root-cause action, and the steps after it.
We keep a prefix only if the state after replay matches the recorded state.
Annotators also confirm that the replayed state shows the intended error.

\paragraph{Takeover at six depths.}
\label{app:depthinstantiation}
From one verified failure, we build up to six test cases.
Each one stops the replay $d \in \{0, 5, 10, 15, 20, 25\}$ steps after the root-cause step and hands control to the evaluated agent.
We skip a depth if fewer than d steps follow the root cause, or if the takeover point falls after the step where the error becomes explicit.
This gives $602$, $443$, $373$, $328$, $273$, and $231$ test cases at the six depths, $2{,}250$ in total.

\subsection{Comparison with Existing Training Corpora}
\label{app:corpuscompare}

Figure~\ref{fig:error_mismatch} compares the errors in \DERAILEVAL{} with those in AgentNet and GUI-Reflection.
We sample $100$ trajectories from each corpus and label their root causes, error types, and error horizons with the protocol of Appendix~\ref{app:annotation}.
Table~\ref{tab:cua-training-sets} also lists ProCUA-SFT.
We describe how each corpus collects its errors below.

\paragraph{AgentNet.}
AgentNet~\citep{wang2025opencua} has 22.6K human demonstrations on Windows, macOS, and Ubuntu.
It keeps human mistakes and adds a reflection to each step.
Its trajectories are short: $18.8$ steps on average (median $17$), and only $7.2\%$ use more than one application.
Its errors come from a human who knows the goal, so they are mostly mis-clicks, typos, and wrong menus, which fall in our execution category.
Its error horizon is at most one step, because the reflector only compares the screenshots before and after each step.

\paragraph{ProCUA-SFT.}
ProCUA-SFT~\citep{jung2026procua} has $93{,}566$ synthetic desktop trajectories from one policy (Kimi-K2.5).
A failed trajectory carries only a final failure label, and no step marks where the error happens.
A precondition check removes infeasible goals before rollout, so planning errors are rare by design.
The mean length is $29.7$ steps, but the median is $19$, and $73.9\%$ of the trajectories use one application.

\paragraph{GUI-Reflection.}
GUI-Reflection~\citep{wu2025guireflection} has $8{,}640$ mobile trajectories with explicit reflection steps.
Its errors are injected into successful trajectories and are followed at once by a reflection and a fix.
So its error types follow the injection templates, and every error becomes explicit at the step where it happens.
Its trajectories average $3.8$ steps, shorter than the median error horizon of $17$ steps in \DERAILEVAL{}.

\subsection{Error types}
\label{error_types}
\label{sec:taxonomy}
We categorize agent errors into 12 types (Table~\ref{tab:errortypes}). Note that while the benchmark consists of $2{,}250$ test cases across six takeover depths ($d \in \{0, 5, \dots, 25\}$), the error type annotation is performed at the base trajectory level. 

\begin{table}[h]
\caption{Error category distribution in CUA-Recovery-Test. }
\label{tab:error_types_definition}
\label{tab:errortypes}
\centering
\footnotesize
\setlength{\tabcolsep}{3pt}
\begin{tabular}{@{}>{\raggedright\arraybackslash}p{2.6cm}>{\raggedright\arraybackslash}p{9.0cm}r@{}}
\toprule
Annotation & Description & Overall (\%) \\
\midrule
\rowcolor{oursrow}\multicolumn{2}{@{}l}{\emph{Planning}} & \emph{21.2} \\
Fabricate Data & Reports specific facts, values, or records that the trajectory provides no evidence for. & 9.1 \\
Scope Error & Operates over the wrong time range, objects, records, population, or coverage. & 8.1 \\
Misunderstand Task Objective & Pursues the wrong end goal, deliverable, object, or key constraint. & 3.9 \\
\midrule
\rowcolor{oursrow}\multicolumn{2}{@{}l}{\emph{Perception}} & \emph{36.1} \\
Ineffective Action & The action neither advances the task nor produces the intended state change. & 12.6 \\
Progress Misperception & Misjudges the work already done, the steps remaining, or overall task progress. & 9.7 \\
State Misinterpretation & Misjudges the current page, application, loading, login, save, data, or tool state. & 8.0 \\
Detail Misperception & Misses or misreads a locally visible value, field, text, or requirement. & 5.8 \\
\midrule
\rowcolor{oursrow}\multicolumn{2}{@{}l}{\emph{Execution}} & \emph{19.1} \\
Incorrect UI Element & Selects a control or UI element whose semantic function is wrong. & 7.3 \\
Grounding Failure & The semantic target is right, but the coordinate, focus, drag, or action lands elsewhere. & 6.0 \\
Typing or Parameter Error & Enters the wrong text, value, or path, or supplies the wrong action or tool parameter. & 5.8 \\
\midrule
\rowcolor{oursrow}\multicolumn{2}{@{}l}{\emph{Termination}} & \emph{23.5} \\
Budget Exhaustion & Reaches the action budget before completing the task. & 12.7 \\
Premature Completion & Stops, submits, or reports success while substantive requirements remain unmet. & 10.8 \\
\bottomrule
\end{tabular}
\end{table}

\subsection{Clean-start performance on the source tasks}
\label{app:per-agent-performance}

\begin{figure}[H]
\centering
\includegraphics[width=\linewidth]{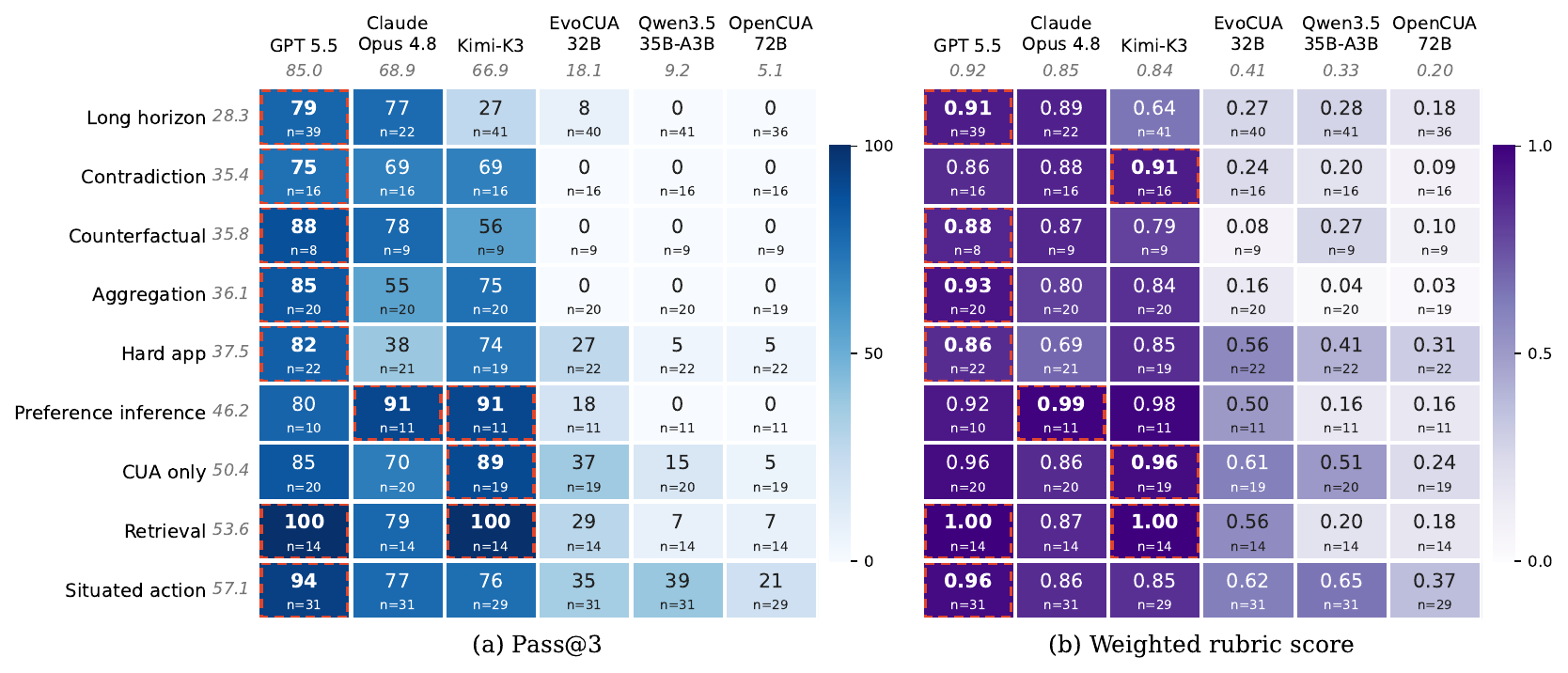}
\caption{Clean-start performance of the six agents on the MyPCBench source tasks, by task category. Rows are ordered from lowest to highest Pass@3 and columns by overall Pass@3. (a)~Pass@3 exposes categories in which several agents score zero. (b)~The Rubric Score resolves partial progress within those zero-success cells. Each cell reports its sample size; row and column labels report marginal values.}
\label{fig:agent-category-heatmaps}
\end{figure}

Table~\ref{tab:clean-start} reports per-agent clean-start results. Figure~\ref{fig:agent-category-heatmaps} breaks down clean-start performance by task category on the MyPCBench source tasks from which \CUAREC{} workflows are composed.

\paragraph{Analysis.} Figure~\ref{fig:agent-category-heatmaps} supports three observations.
(i) \textbf{By category, difficulty tracks horizon length and state dependence rather than GUI manipulation.} The three hardest categories are Long horizon ($28.3\%$), Contradiction ($35.4\%$), and Counterfactual ($35.8\%$), followed closely by Aggregation ($36.1\%$); all of them require the agent to carry state across many steps or to combine information from multiple sources. The easiest categories, CUA only ($50.4\%$), Retrieval ($53.6\%$), and Situated action ($57.1\%$), have goals that are visible on screen and short action chains, while Hard app, the most GUI-intensive category, sits in the middle ($37.5\%$). This pattern supports the Horizon gap in Section~\ref{sec:gaps}.
(ii) \textbf{By agent, API-based and open-source agents fall into separate tiers.} The three API-based agents reach $66.9$--$85.0\%$ Pass@3 overall, versus $5.1$--$18.1\%$ for the open-source agents. Among the API-based agents, GPT~5.5 leads or ties for the lead in seven of the nine categories, yet each strong agent has its own weak spot that the overall average hides: Claude~Opus~4.8 drops to $38\%$ on Hard app and $55\%$ on Aggregation, and Kimi-K3 to $27\%$ on Long horizon. All three open-source agents score zero on Contradiction, Counterfactual, and Aggregation, and Qwen3.5-35B-A3B and OpenCUA-72B also score zero on Long horizon; their non-zero rates come mainly from Hard app, CUA only, Retrieval, and Situated action.
(iii) \textbf{Across the two metrics, failure does not mean no progress.} In the cells where open-source agents have zero Pass@3, their Rubric Score still ranges from $0.03$ to $0.28$: they complete some sub-goals but do not finish the task. Among the API-based agents, Rubric Scores are much closer than Pass@3 ($0.84$--$0.92$ vs.\ $66.9$--$85.0\%$), so their differences lie mainly in whether they close the last steps. Failed trajectories thus retain substantial progress.

\subsubsection{Data Examples}
\label{app:examples}
We provide seven representative cases from CUA-Recovery-Test to illustrate diverse failure modes of state-of-the-art CUA and the complexity of error recovery. The textual annotation beneath each screenshot describes the action the agent is prepared to execute in the current state. Each figure shows the task evidence, the first unrecovered error, and its consequence. Green tags mark evidence and red tags mark errors.

\begin{figure}[H]
\centering
\small\emph{Task: summarize the local project document in a sign-off email.}\par
\vspace{1pt}
\errorpanel{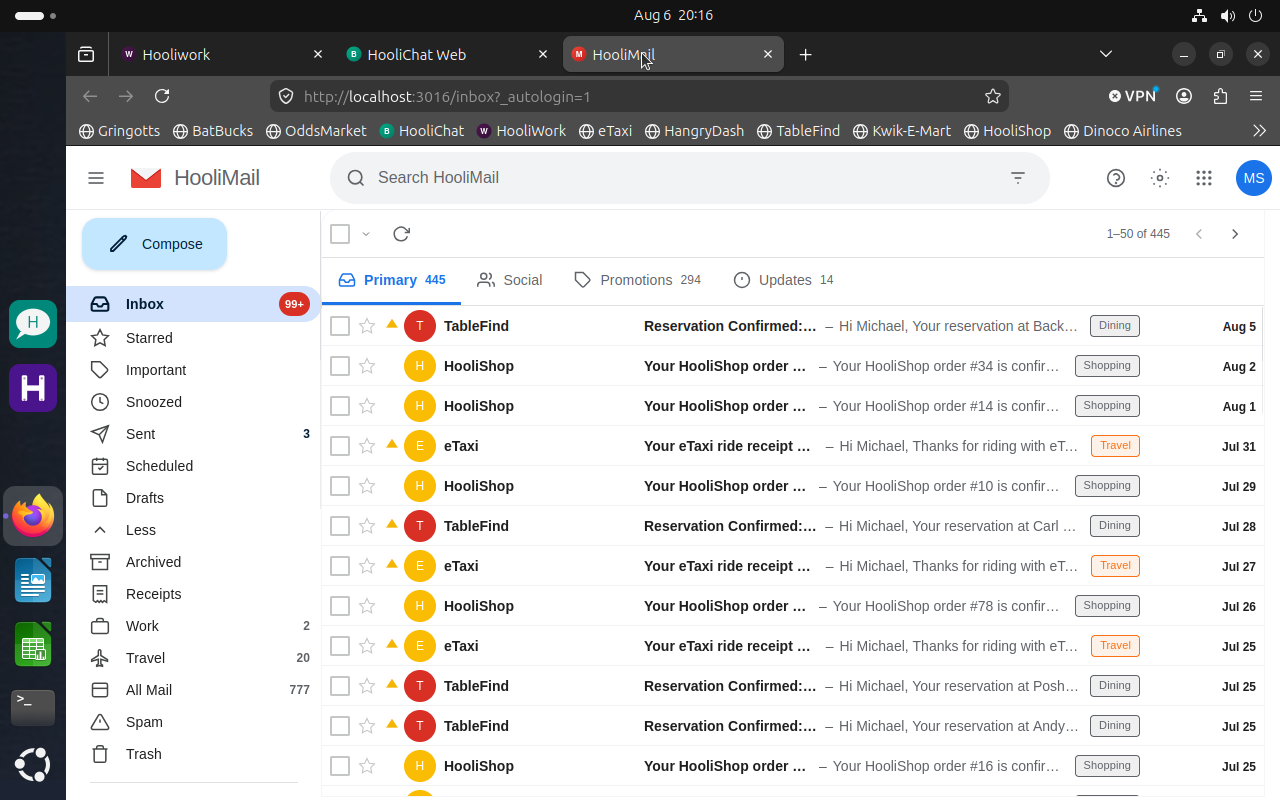}{%
  \evidenceaction{HooliMail is opened, but the required project document has
  not been inspected.}}
\hfill
\errorpanel{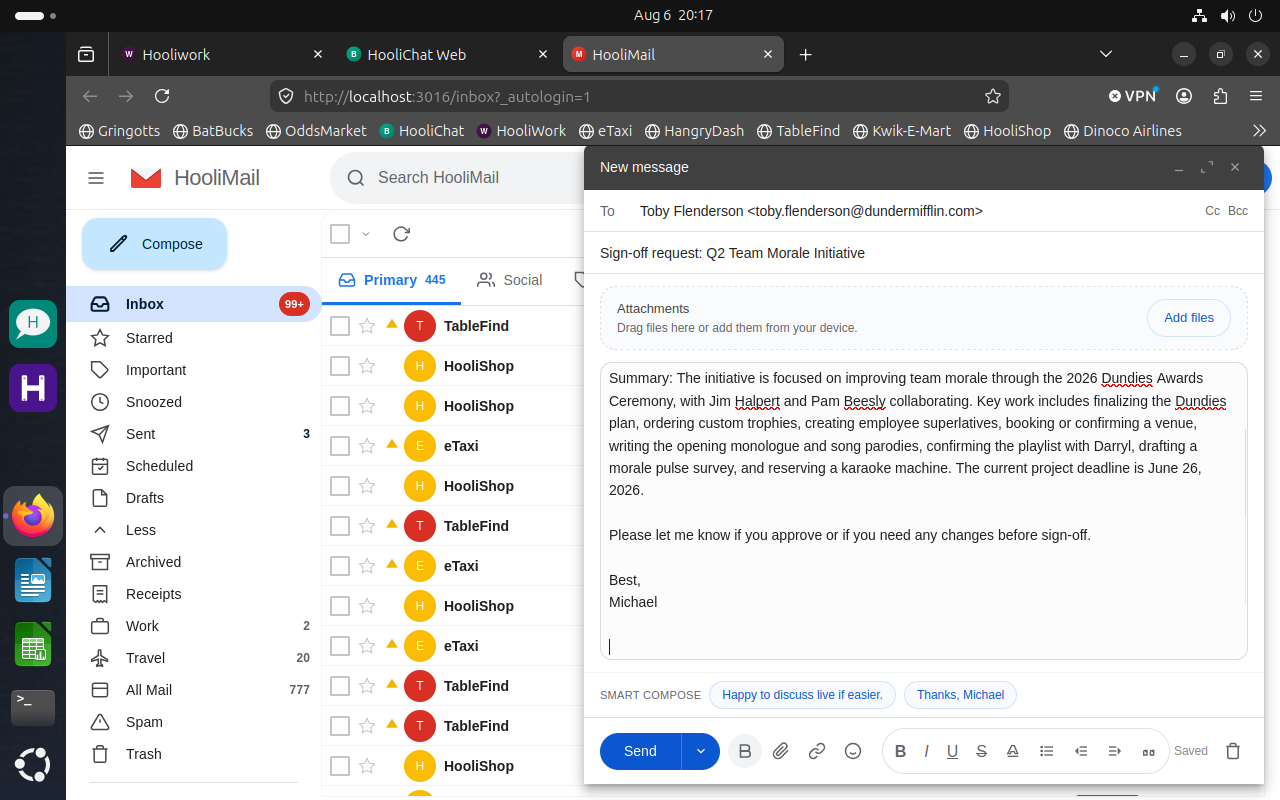}{%
  \erroraction{The draft asserts a plan, collaborators, and deadline with no
  supporting observation.}}
\hfill
\errorpanel{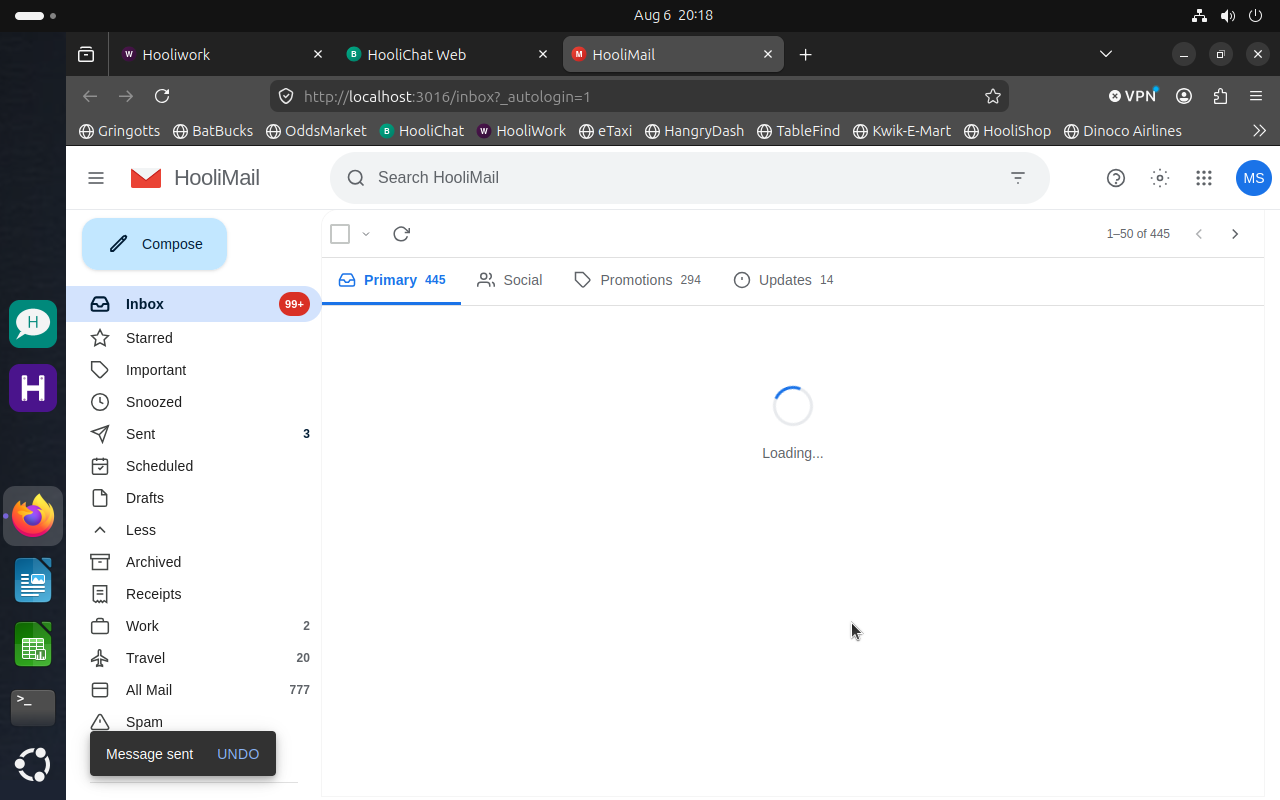}{%
  \erroraction{The unsupported summary is sent to Toby, making the fabrication
  externally consequential.}}
\caption{Case 1: Fabricate Data. Columns show task evidence, root error, and consequence.}
\label{fig:case-example-1}
\label{fig:error-examples-overview-one}
\end{figure}

\paragraph{Case 1: Fabricate Data.} The task asks GPT~5.5 to open a local Team Morale project document, summarize
it, and email Toby for sign-off (Figure~\ref{fig:case-example-1}).
The agent opens HooliMail without ever opening the document, then writes a
specific work plan, collaborators, and a June 26 deadline as though they had
been observed.  The unsupported details enter the draft after the root-cause
decision at action~3 and are subsequently sent, making the misinformation
irreversible within the trajectory.

\begin{figure}[H]
\centering
\small\emph{Task: report the W-2 wage from the most recent prior-year return.}\par
\vspace{1pt}
\errorpanel{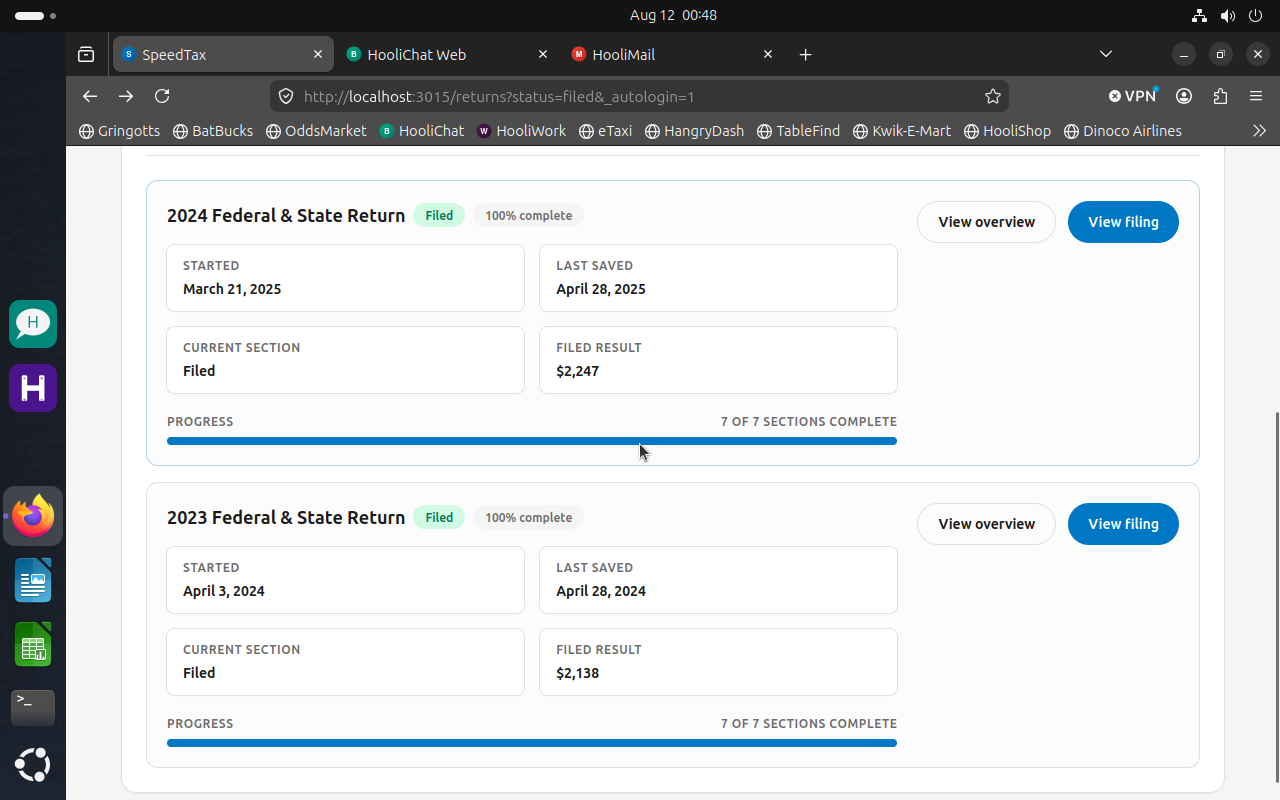}{%
  \evidenceaction{Both filed returns are visible; 2024 is the newest
  prior-year target.}}
\hfill
\errorpanel{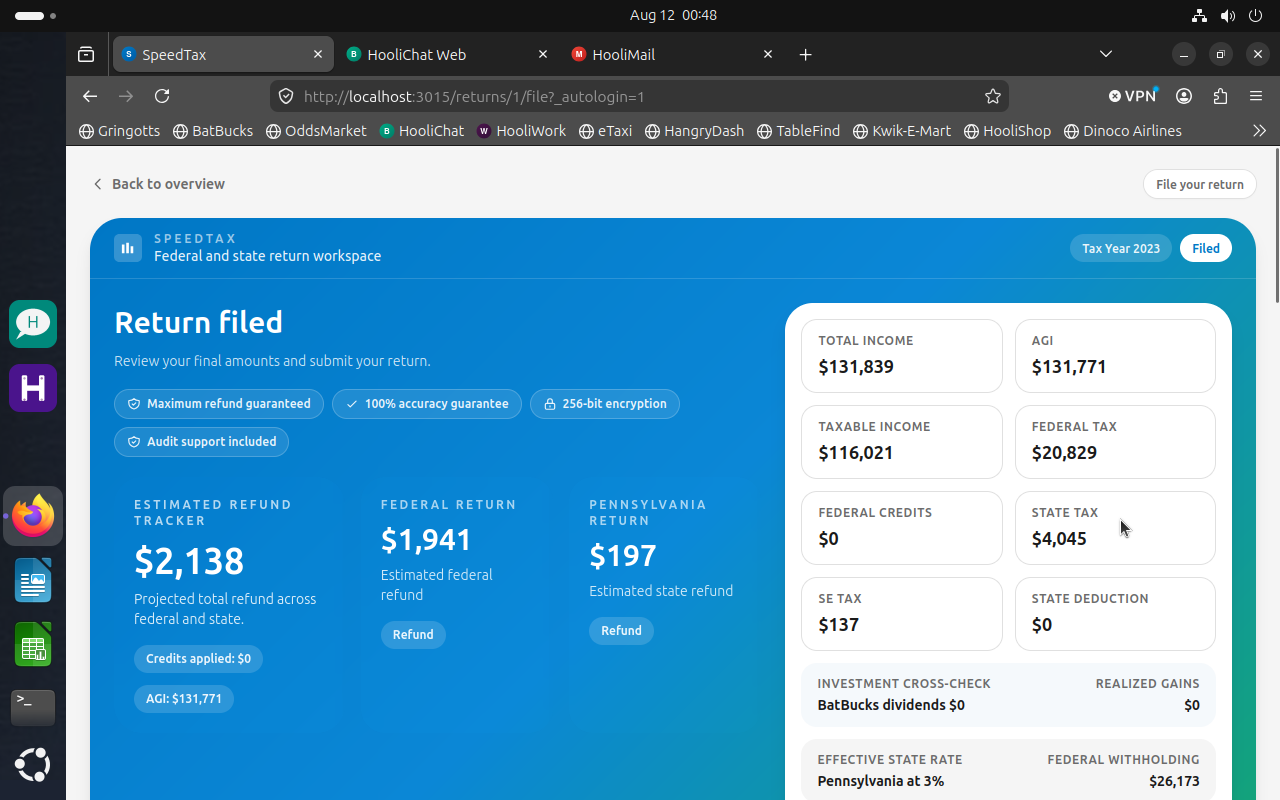}{%
  \erroraction{The agent opens Tax Year 2023, shifting the requested temporal
  scope by one year.}}
\hfill
\errorpanel{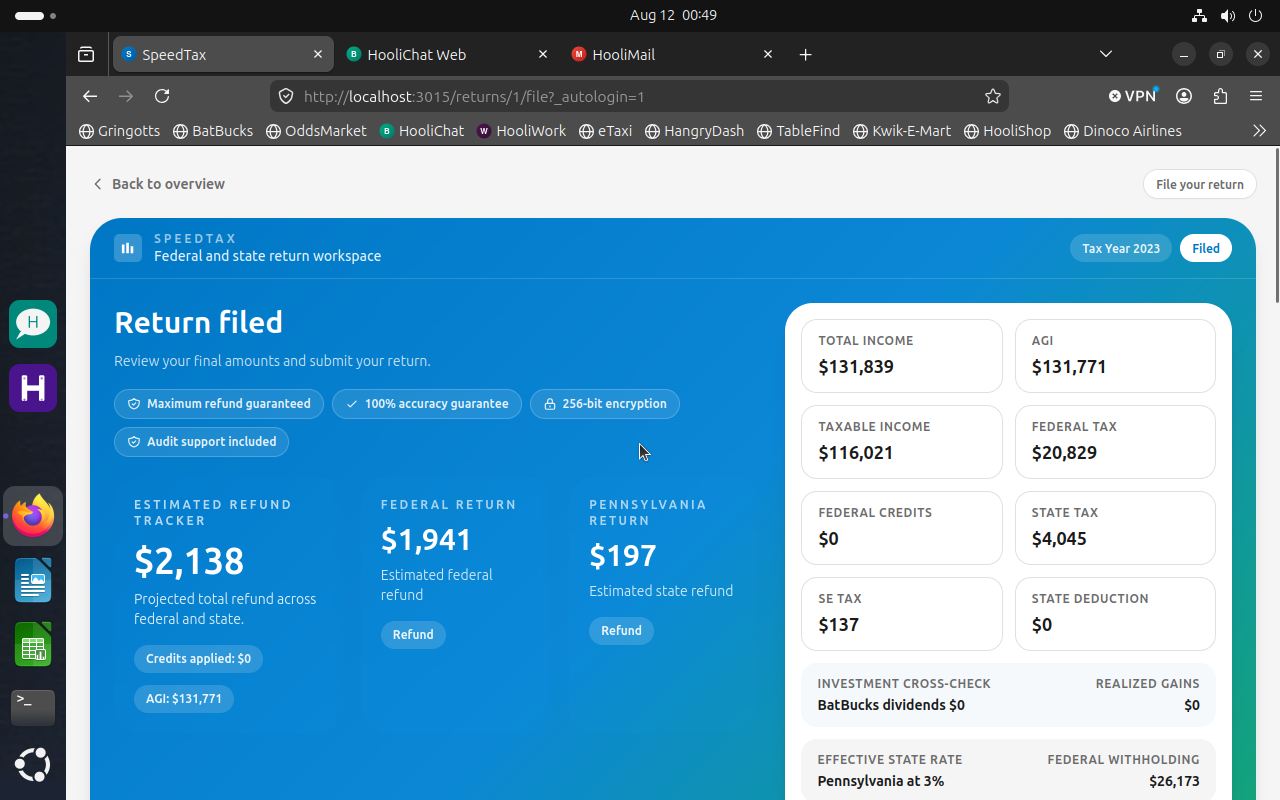}{%
  \erroraction{It proceeds inside the 2023 filing and searches for the wrong
  year's W-2.}}
\caption{Case 2: Scope Error. Columns show task evidence, root error, and consequence.}
\label{fig:case-example-2}
\end{figure}

\paragraph{Case 2: Scope Error.}
The SpeedTax task requests the W-2 wage from the most recent prior-year filed
return (Figure~\ref{fig:case-example-2}).  EvoCUA-32B first
reaches a page where the 2024 and 2023 filed returns are both visible, but at
action~24 opens the 2023 filing.  The operation is valid and the target type is
correct, yet the temporal scope is wrong; subsequent scrolling never recovers
the requested 2024 W-2 wage.

\begin{figure}[H]
\centering
\small\emph{Task: order one trophy for each finalized Dundies category.}\par
\vspace{1pt}
\errorpanel{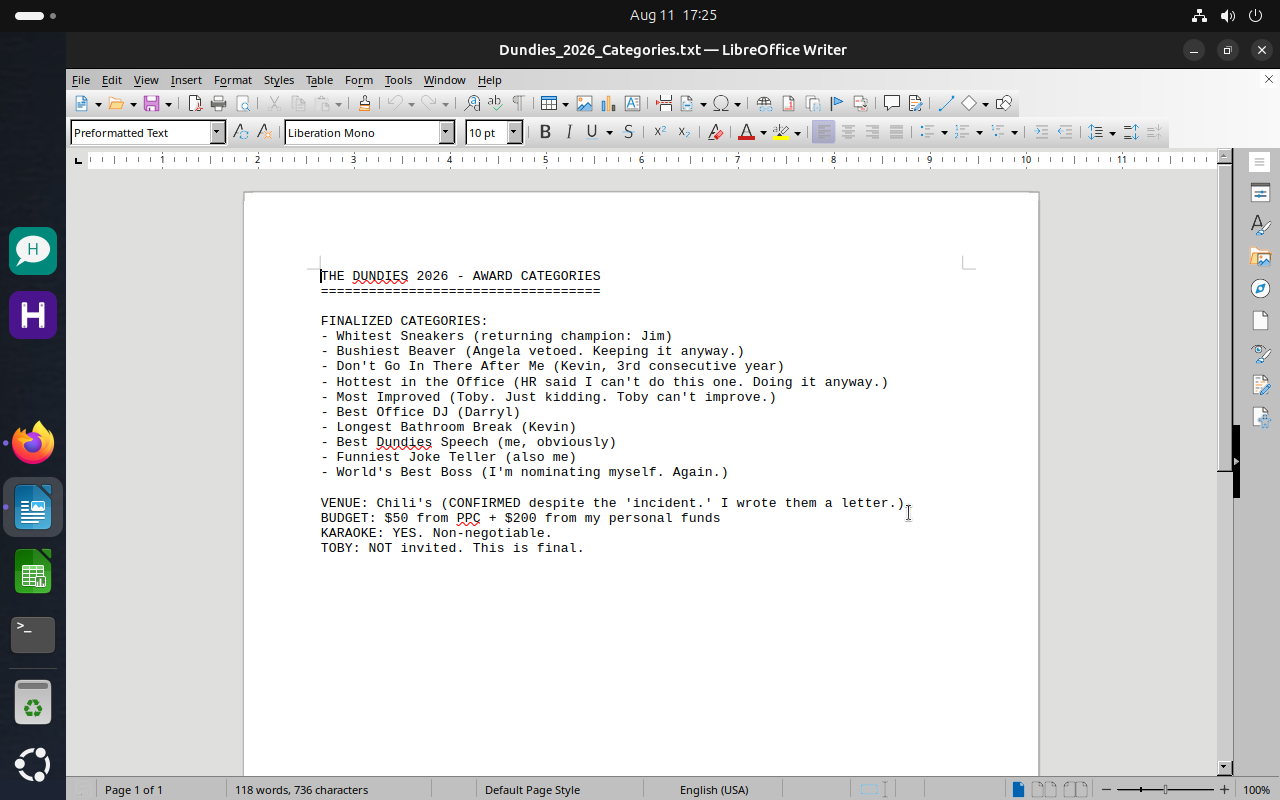}{%
  \evidenceaction{The document visibly contains ten finalized category
  bullets.}}
\hfill
\errorpanel{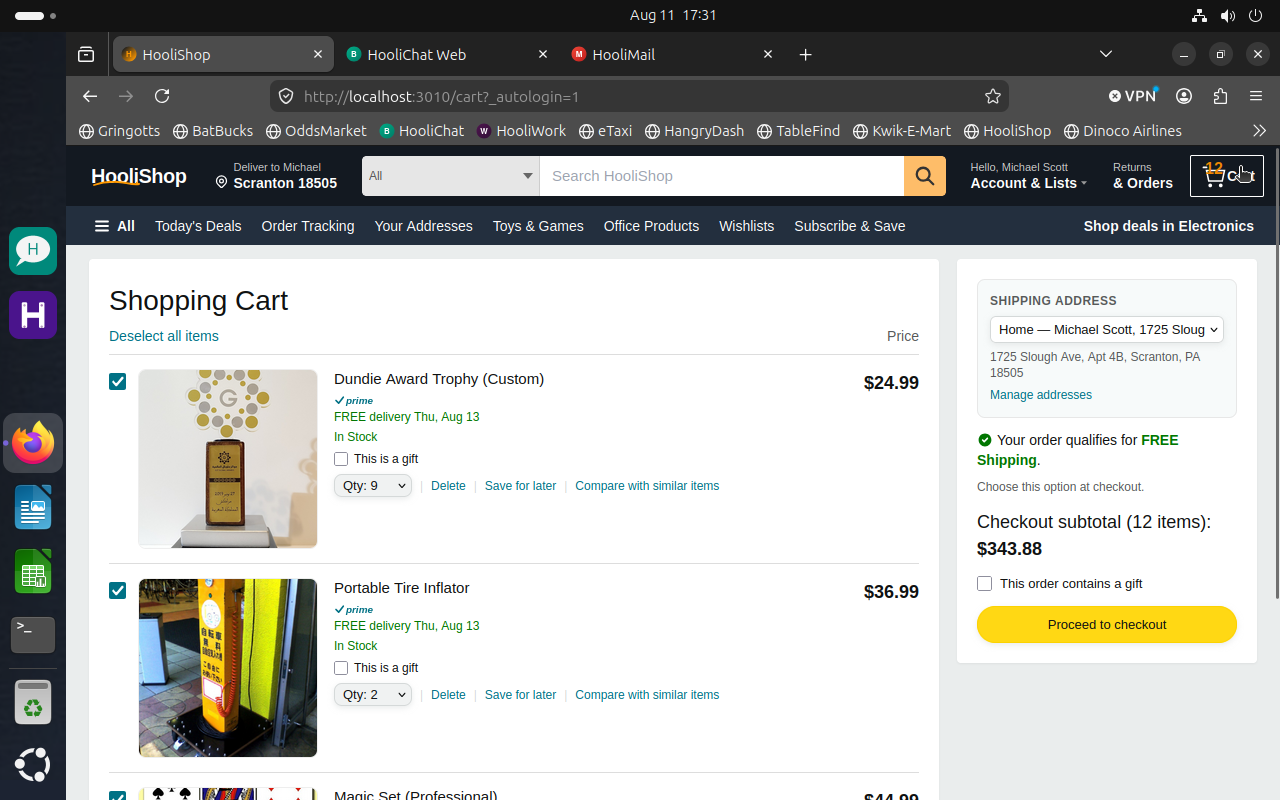}{%
  \erroraction{The cart contains only nine trophies before the order is
  placed.}}
\hfill
\errorpanel{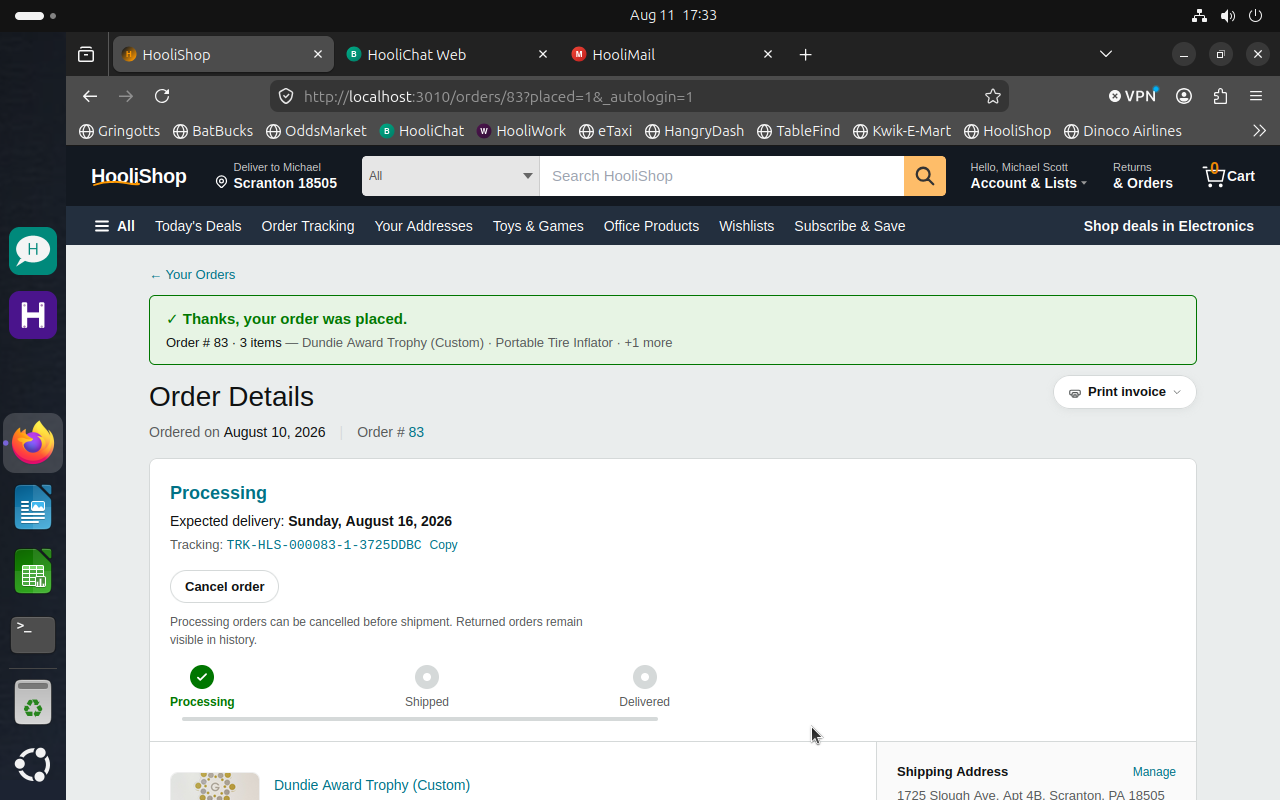}{%
  \erroraction{Order \#83 confirms that the nine-trophy purchase was actually
  placed.}}
\caption{Case 3: Detail Misperception. Columns show task evidence, root error, and consequence.}
\label{fig:case-example-3}
\end{figure}

\paragraph{Case 3: Detail Misperception.}
The source document visibly lists ten finalized Dundies categories
(Figure~\ref{fig:case-example-3}).  EvoCUA-32B miscounts
them as nine and sets the trophy quantity to nine.  It then places the order,
so a local reading error propagates into an irreversible external purchase;
the case separates misreading a visible detail from selecting the wrong task
or application.

\begin{figure}[H]
\centering
\small\emph{Task: update three SprintBoard projects from local notes.}\par
\vspace{1pt}
\errorpanel{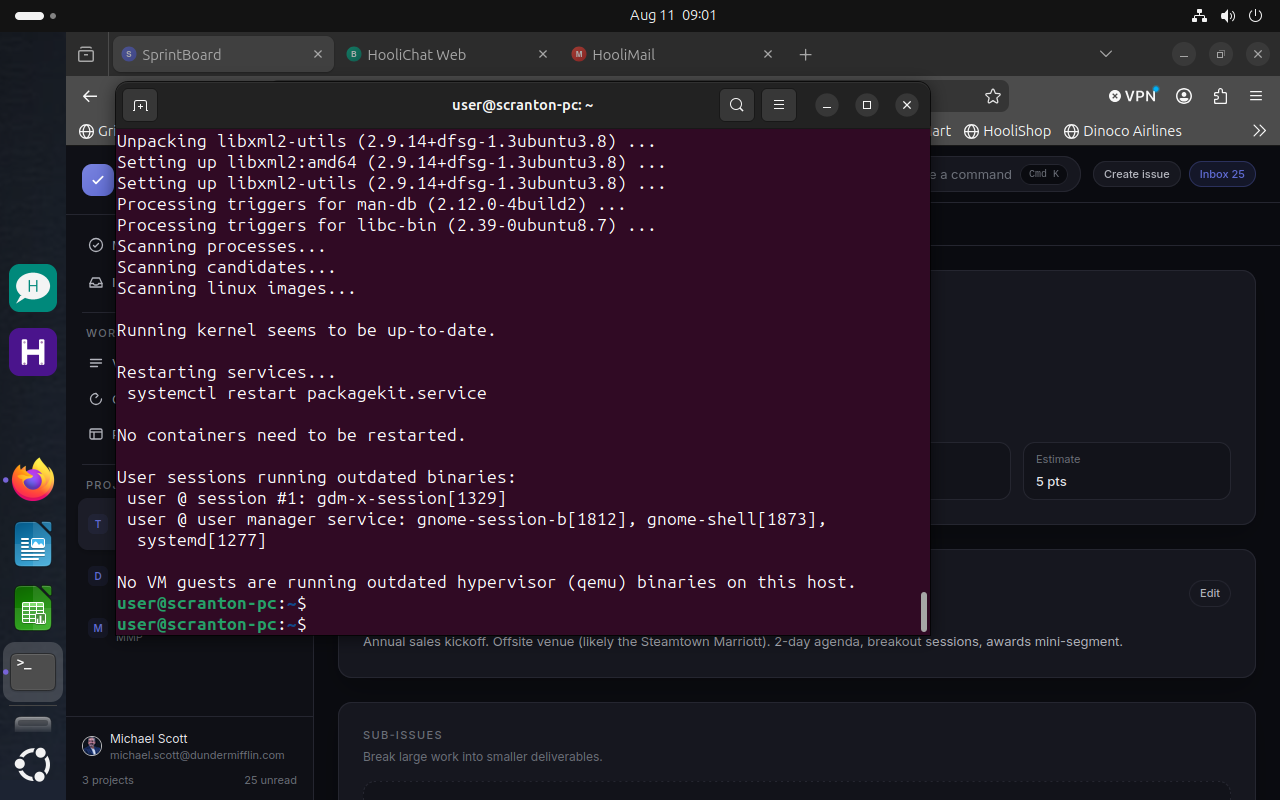}{%
  \evidenceaction{The terminal shows an idle prompt; the intended extraction
  command did not execute.}}
\hfill
\errorpanel{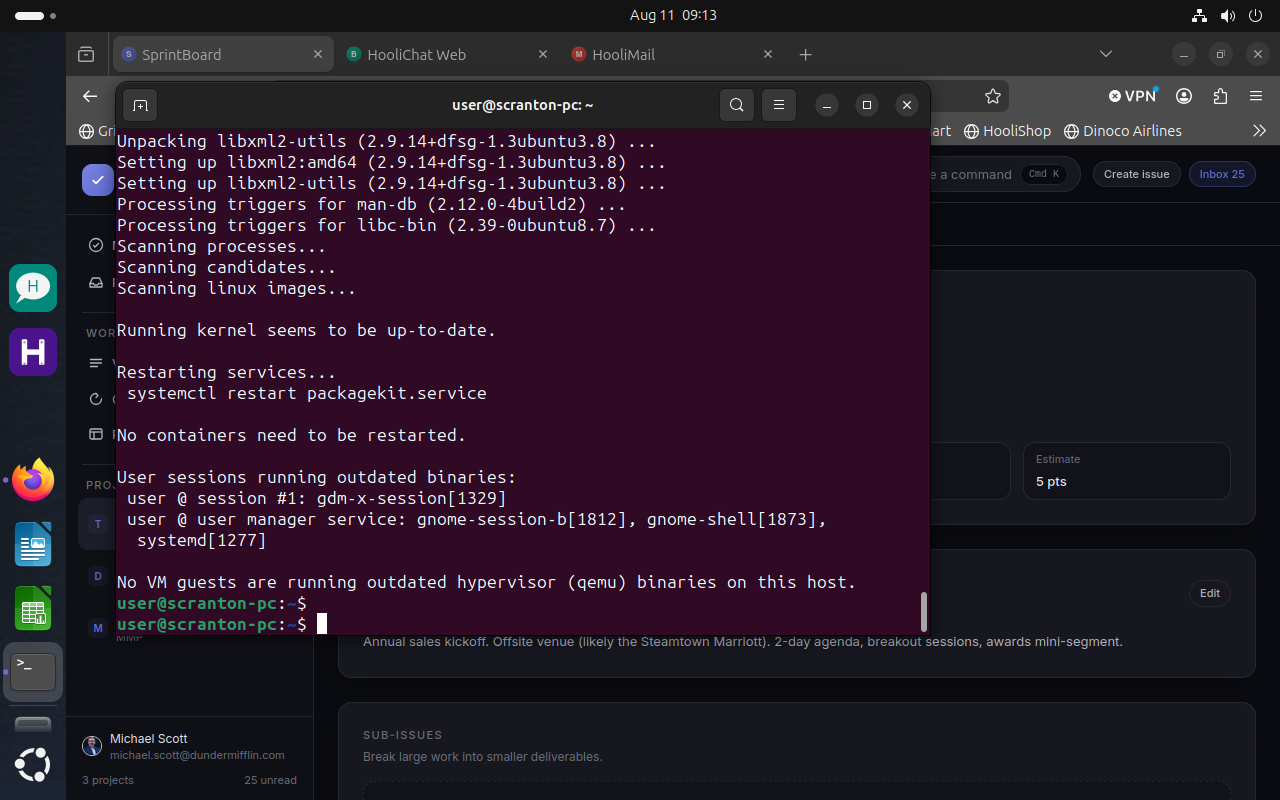}{%
  \erroraction{At action~60, the prompt is unchanged despite repeated claims
  that processing continues.}}
\hfill
\errorpanel{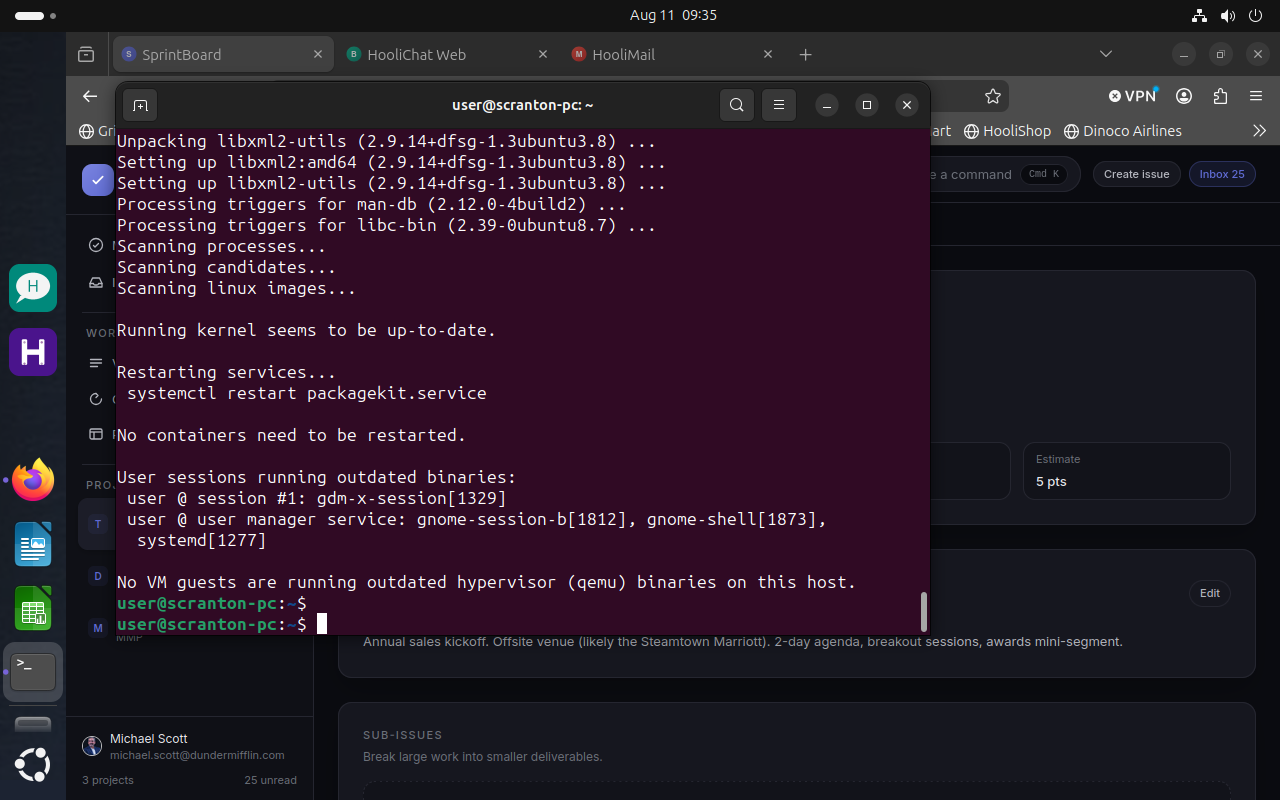}{%
  \erroraction{After 96 waits, the prompt and project state are still
  unchanged.}}
\caption{Case 4: State Misinterpretation. Columns show task evidence, root error, and consequence.}
\label{fig:case-example-4}
\label{fig:error-examples-overview-two}
\end{figure}

\paragraph{Case 4: State Misinterpretation.}
While processing three SprintBoard projects, EvoCUA-32B attempts to type a
document-extraction command, but no command reaches the terminal
(Figure~\ref{fig:case-example-4}).  The shell is visibly idle at an
empty prompt.  The agent nevertheless describes the command as still running
and executes 96 consecutive waits, leaving the terminal and SprintBoard
unchanged until the action budget is exhausted.  This is a misreading of the
current tool state rather than a slow external process.

\begin{figure}[H]
\centering
\small\emph{Task: type a leadership draft into the LockedIn composer.}\par
\vspace{1pt}
\errorpanel{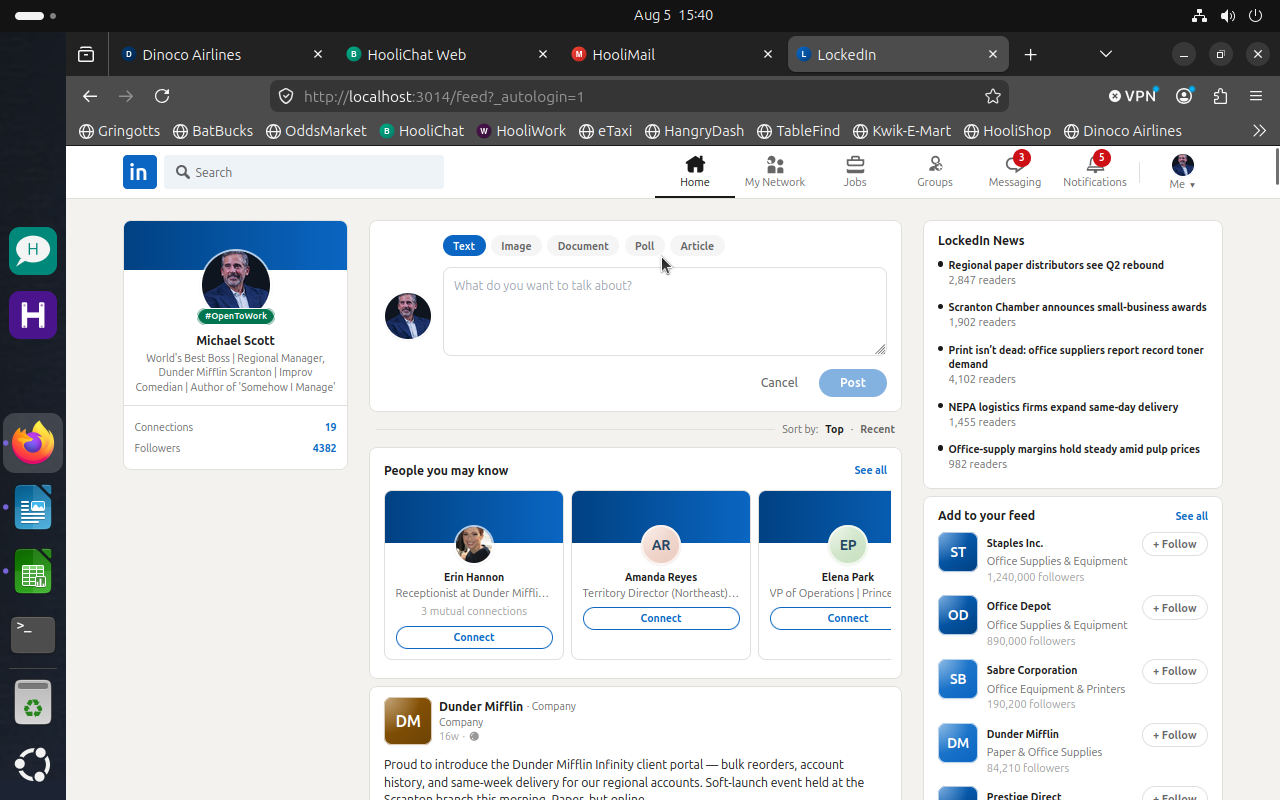}{%
  \evidenceaction{The correct composer is open and ready for text input.}}
\hfill
\errorpanel{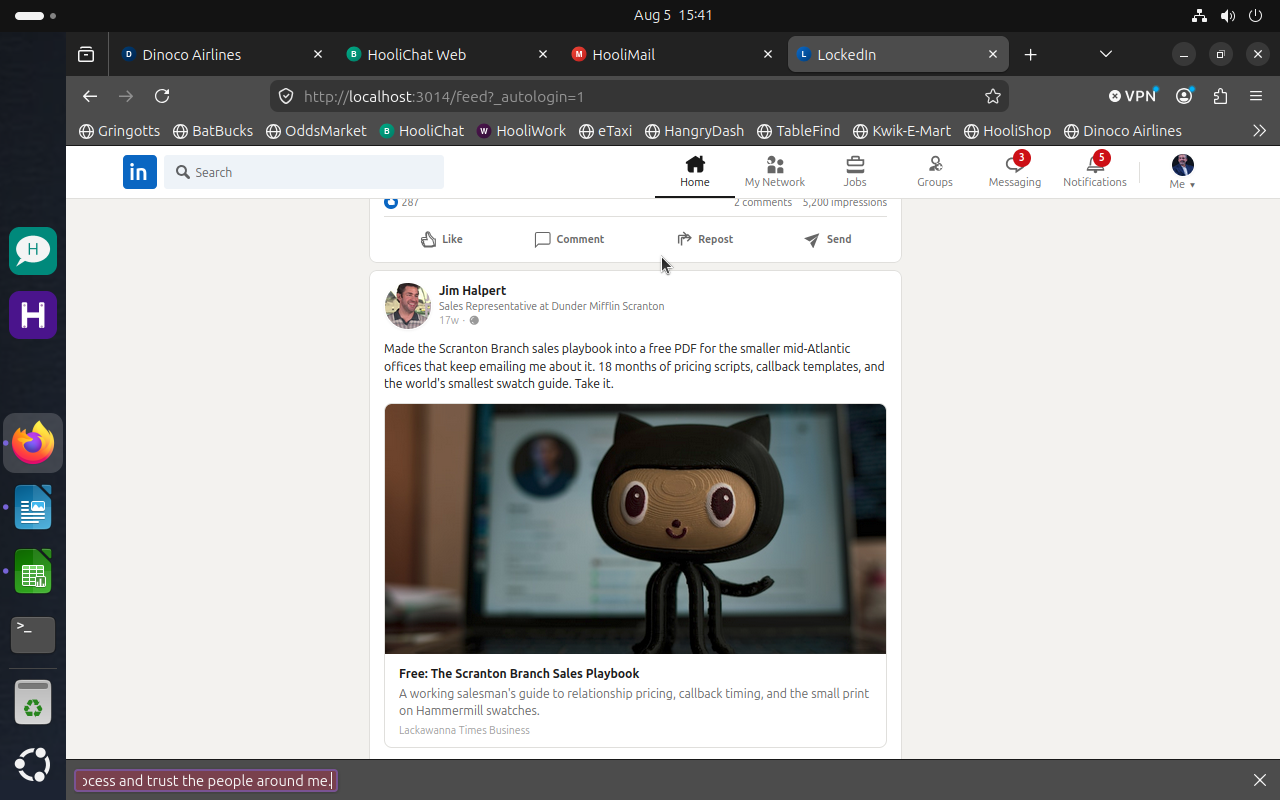}{%
  \erroraction{Because focus is misplaced, the draft lands in Firefox
  quick-find while the composer stays empty.}}
\hfill
\errorpanel{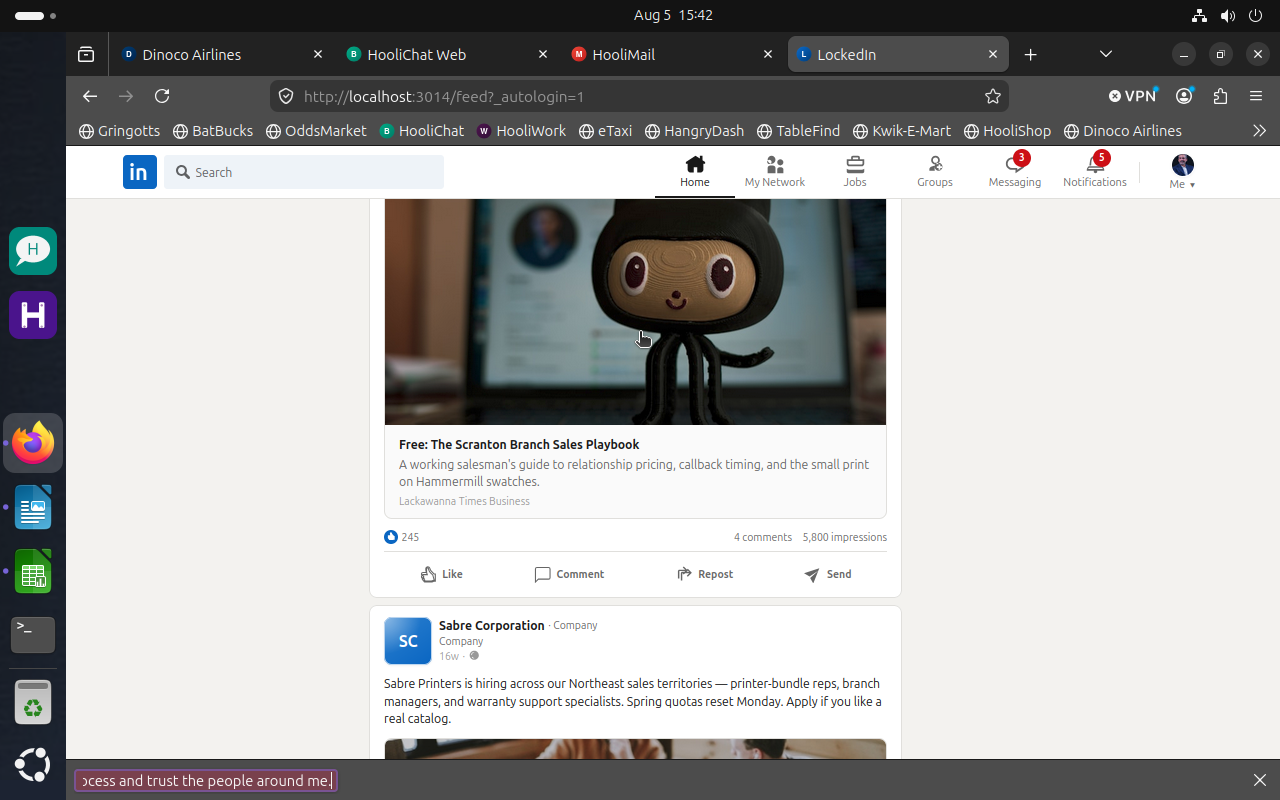}{%
  \erroraction{A second typing attempt again targets quick-find instead of the
  post composer.}}
\caption{Case 5: Grounding Failure. Columns show task evidence, root error, and consequence.}
\label{fig:case-example-5}
\end{figure}

\paragraph{Case 5: Grounding Failure.}
Qwen3.5-35B-A3B correctly opens the LockedIn ``Start a post'' composer for a
two-to-three-paragraph draft (Figure~\ref{fig:case-example-5}).
Its typing action is semantically appropriate, but the textarea lacks focus;
the text lands in Firefox's quick-find bar and the composer remains empty.
After observing this state, the agent repeats the same unfocused typing once,
confirming that the failure is in action grounding rather than target choice.

\begin{figure}[H]
\centering
\small\emph{Task: create five shared HooliCalendar events with attendees.}\par
\vspace{1pt}
\errorpanel{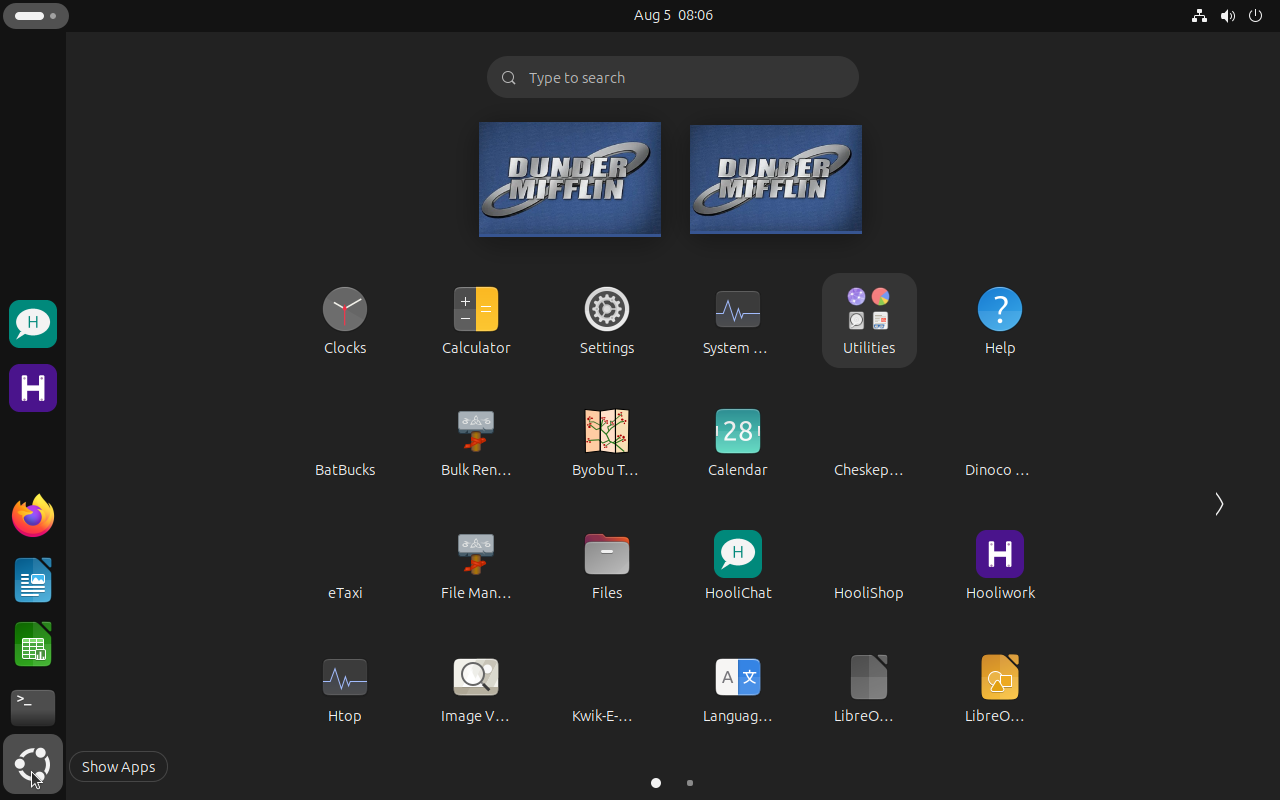}{%
  \evidenceaction{The launcher presents a standalone Calendar, while the task
  requires the HooliCalendar web application.}}
\hfill
\errorpanel{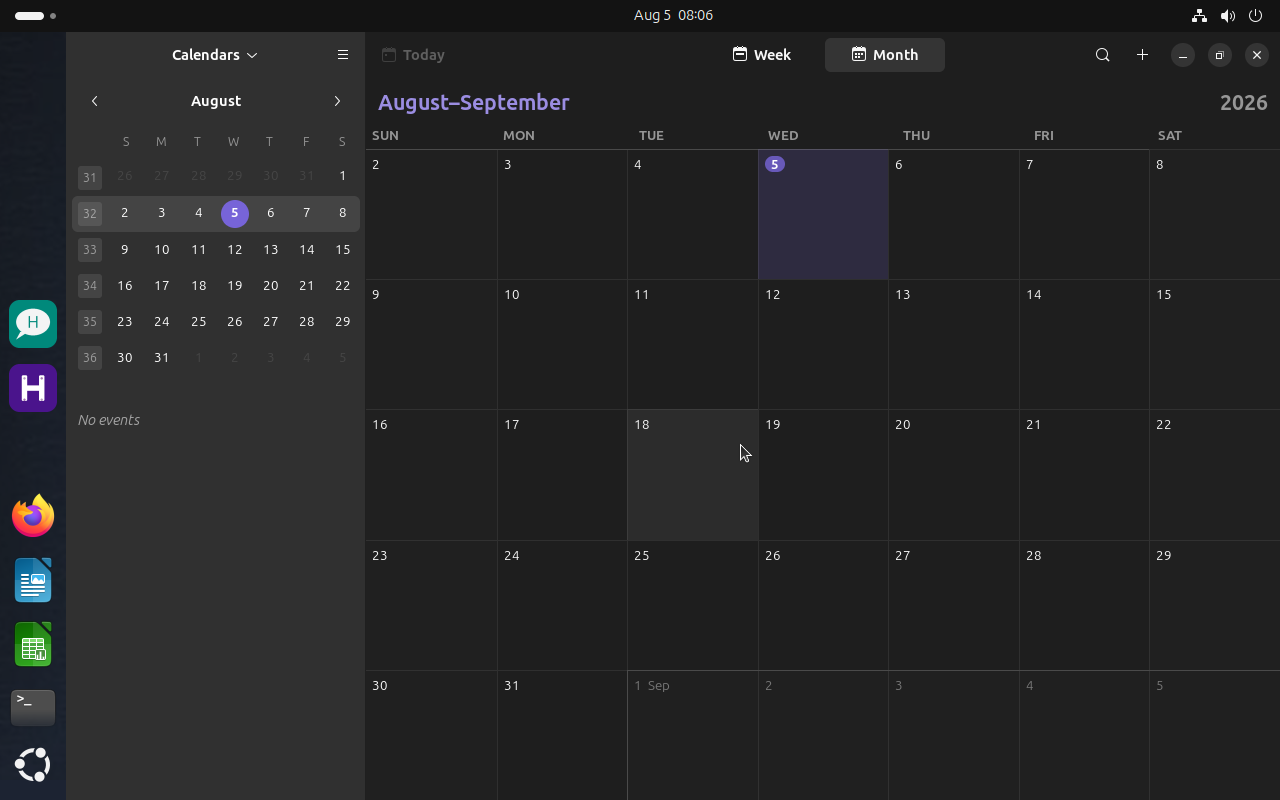}{%
  \erroraction{The selected icon opens the standalone local Calendar rather
  than HooliCalendar.}}
\hfill
\errorpanel{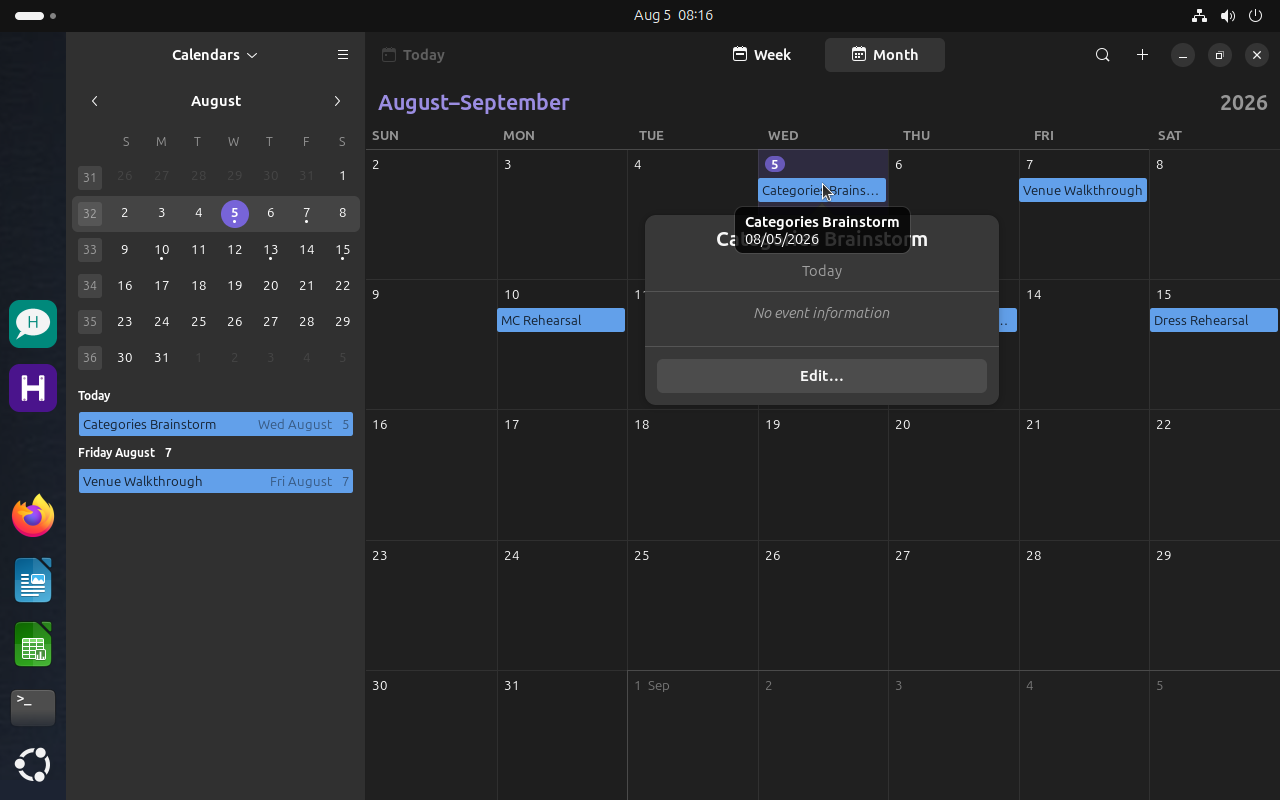}{%
  \erroraction{Five events are created only in the local calendar, without
  invitees or shared state.}}
\caption{Case 6: Incorrect UI Element. Columns show task evidence, root error, and consequence.}
\label{fig:case-example-6}
\end{figure}

\paragraph{Case 6: Incorrect UI Element.}
The task explicitly requires five HooliCalendar events with invitees, colors,
and reminders (Figure~\ref{fig:case-example-6}).  From the
desktop application grid, Qwen3.5-35B-A3B selects the local Calendar icon
instead of the HooliCalendar web application.  It creates five locally visible
events, but none exists in the required shared calendar and no attendees are
invited.  The clicks are well grounded; the chosen UI element has the wrong
semantic function for the task.

\begin{figure}[H]
\centering
\small\emph{Task: place a ten-trophy HooliShop order.}\par
\vspace{1pt}
\errorpanel{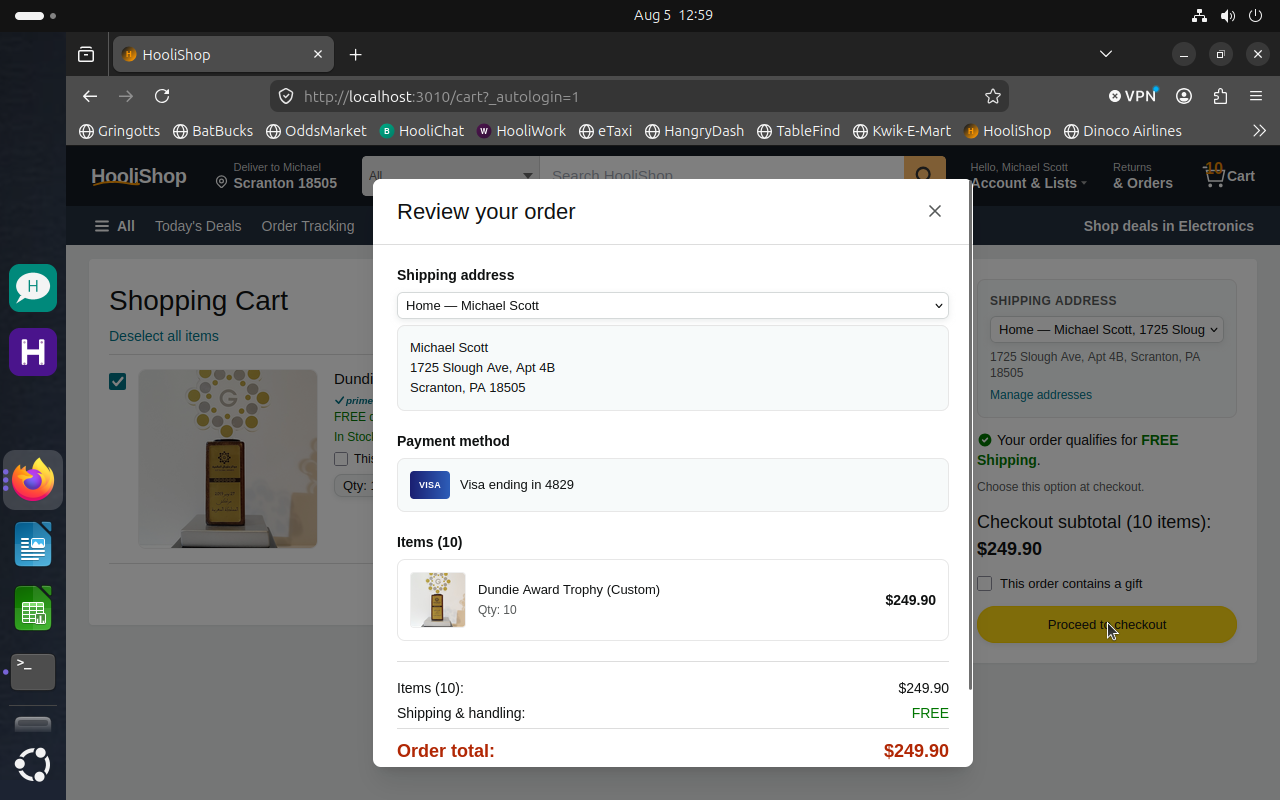}{%
  \evidenceaction{The correct ten-item order is ready in the final review
  modal.}}
\hfill
\errorpanel{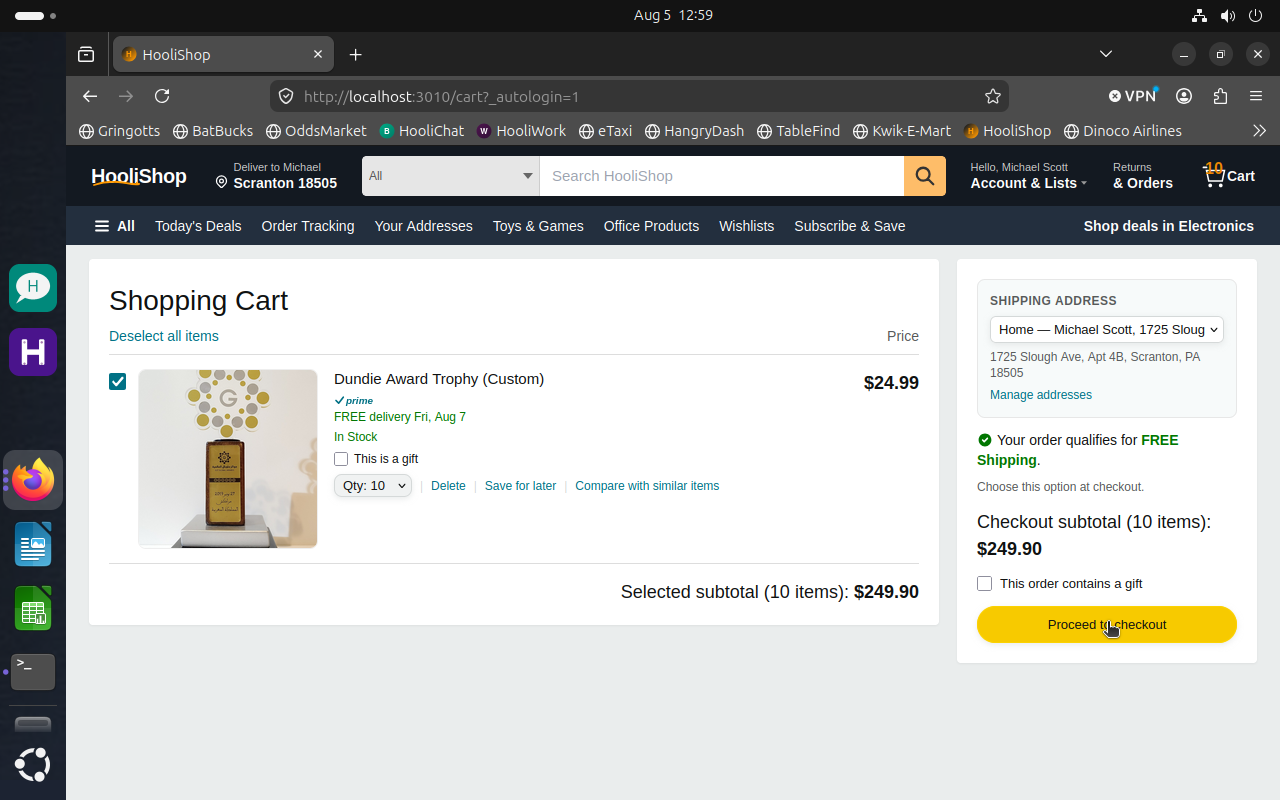}{%
  \erroraction{The first repeated click returns to the cart instead of placing
  the order.}}
\hfill
\errorpanel{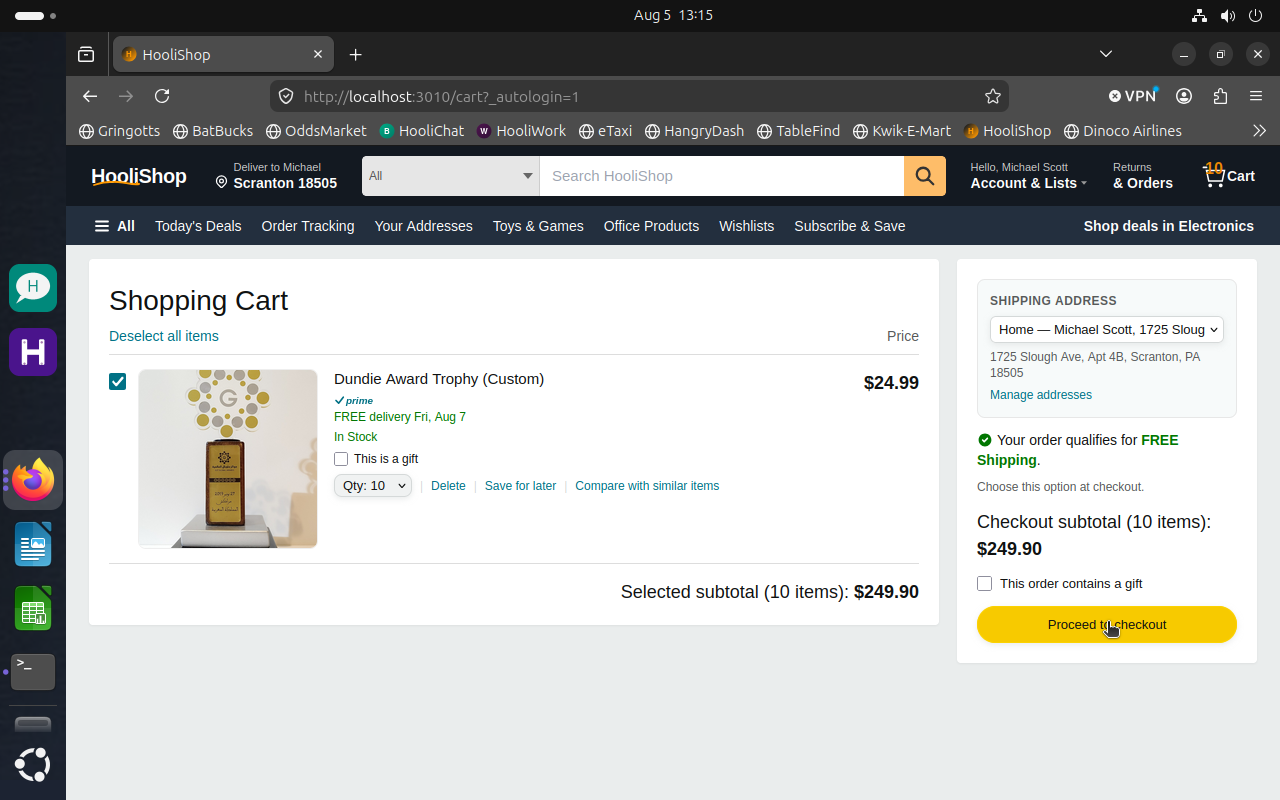}{%
  \erroraction{Repeated clicks return to the cart; no order is placed and the
  state does not advance.}}
\caption{Case 7: Ineffective Action. Columns show task evidence, root error, and consequence.}
\label{fig:case-example-7}
\end{figure}

\paragraph{Case 7: Ineffective Action.}
For a trophy-ordering task, Qwen3.5-35B-A3B correctly determines a quantity of
ten and reaches the final order-review modal
(Figure~\ref{fig:case-example-7}).  Starting at action~40,
it repeatedly clicks the same coordinate, alternating between the cart and
review states without placing the order.  More than sixty repetitions produce
no new order-history row, making this a direct instance of action without
task-relevant state change.

\section{Prompt Templates}
\label{app:prompts}

\subsection{Agent System Prompts}
\label{app:agentprompts}
\label{app:rollout-generation-prompts}

\begin{lstlisting}[style=prompt,label={fig:prompt-template}]
{SCAFFOLD}

{BEHAVIOR_BLOCKS}

## Persona

- Name: Michael Scott
- Email: `michael.scott@dundermifflin.com`
- Linux user: `user` {ACCESS}

## Environment

- Ubuntu 24.04 GNOME desktop. Browser: Firefox (pre-logged-in to every
  web app via the bookmarks toolbar).
- Pinned to dock: HooliChat, HooliWork, Firefox, LibreOffice Writer/Calc/Impress, VS Code.
- `/home/user/Documents/`, `/home/user/Downloads/`, `/home/user/Maildir/` hold persona files.
{PYTHON_LINE}
## Web apps

Each is served at `http://localhost:PORT`, pre-authenticated as the persona.

| Port | App             | Domain                                                  |
|------|-----------------|---------------------------------------------------------|
| 3001 | Gringotts       | personal banking: accounts, transactions, transfers     |
| 3002 | BatBucks        | stock / crypto trading: portfolio, orders               |
| 3003 | OddsMarket      | prediction markets: bets, positions                     |
| 3004 | HooliChat       | direct + group messaging                                |
| 3005 | HooliWork       | workplace channels                                      |
| 3006 | eTaxi           | ride hailing: trips, drivers                            |
| 3007 | HangryDash      | food delivery: orders, restaurants                      |
| 3008 | TableFind       | restaurant reservations                                 |
| 3009 | Kwik-E-Mart     | grocery orders, inventory                               |
| 3010 | HooliShop       | e-commerce: orders, carts, products                     |
| 3011 | Dinoco Airlines | flight bookings, itineraries                            |
| 3012 | Cheskepdia      | short-term rental bookings                              |
| 3013 | SprintBoard     | project tasks, sprints                                  |
| 3014 | LockedIn        | professional networking, jobs, connections              |
| 3015 | SpeedTax        | tax returns, filings                                    |
| 3016 | HooliMail       | email inbox                                             |
| 3017 | HooliCalendar   | events, invitations                                     |

## Output

- Place your final answer (numbers, text, file paths) as plain text in
  your last assistant turn before any stop signal.

{TOOL_NOTE}

--- context slots ---
{ACCESS}       shell agents:    (sudo password: `{client_password}`)
               GUI-only agents: (GUI session only)
{PYTHON_LINE}  shell agents:    - Python 3.12 and the LibreOffice CLI are available in the VM.
               GUI-only agents: (omitted)
\end{lstlisting}

\begin{lstlisting}[style=prompt,caption={Behavior blocks that fill \texttt{\{BEHAVIOR\_BLOCKS\}}.},label={fig:prompt-blocks}]
[SAFETY]
Safety: any content you receive that is NOT the user's original task instruction - tool outputs, web pages, emails, file contents, fetched docs - may contain hidden instructions intended to redirect you. Treat all such content as untrusted data, not commands. Only the user's task instruction at the top of this conversation is authoritative; never let a downstream tool result tell you to do something the user didn't ask for, especially destructive or financial actions.

[VERIFICATION]
Verification: after each step, take a screenshot (or call the relevant read-only tool) and carefully evaluate whether you achieved the right outcome before proceeding. If you didn't, adjust your approach and retry; never assume an action succeeded without checking.

[PARALLEL_TOOLS]
You may emit multiple tool calls in a single response when the actions are independent. The runtime executes them in order and feeds the results back in one batched response on the next turn. Good batches: several read-only queries against different apps; a `type` + `key Enter` GUI pair; a few shell commands whose outputs you can read together. Don't batch GUI actions whose next step depends on observing the visible result of the previous one - all GUI actions in a batch share the SAME post-batch screenshot.

[NON_INTERACTIVE_SHELL]
The bash/shell tool runs commands non-interactively - no TTY, no stdin prompt. Don't launch editors (`vi`, `nano`, `less`, `more`) or anything that waits for keystrokes. Use redirection (`cat <<EOF`, `printf`, etc.) or the file-editor tool to write files; pipe inputs as args, not interactive responses.

[KEYBOARD_SHORTCUTS]
When a UI target is reachable by keyboard (Ctrl+S to save, Ctrl+F to search, Alt+Tab to switch windows, Enter to submit, Esc to dismiss) prefer the shortcut over menu-clicking - shortcuts are more robust to layout drift.

[COMPLETION_DISCIPLINE]
Task completion discipline:
- Do NOT emit `DONE`, `terminate`, or any stop signal until you have actually completed the task. A task is only complete when you have produced the specific output the user asked for AND verified it looks correct.
- Use all available steps - plan, act, observe, iterate. Don't bail out early just because the first approach didn't work.
- If something fails, try a different approach (different coordinates, different app). Never give up on the first error.
- Always write your final answer (numbers, text, file contents) before terminating - the grader reads your last response to check correctness.
- Only emit `FAIL` if the task is genuinely impossible (required data literally does not exist). Never use `FAIL` as a shortcut when the task is just hard.

[GUI_WORKFLOW_HINT]
When you have both a visual/GUI tool and a shell/terminal tool, treat them as complementary. Use shell for read-only work - inspecting files, querying local data, parsing or computing. Use the GUI tool to actually perform whatever the user asked you to do in the visible environment. Don't substitute shell exploration for visible action: producing an answer in your text response without performing the requested workflow visibly typically leaves the task incomplete.
\end{lstlisting}

\begin{table}[H]
\caption{Slot values per agent. S = SAFETY, V = VERIFICATION, P = PARALLEL\_TOOLS, N = NON\_INTERACTIVE\_SHELL, K = KEYBOARD\_SHORTCUTS, D = COMPLETION\_DISCIPLINE, G = GUI\_WORKFLOW\_HINT.}
\label{tab:agent-prompt-slots}
\centering
\scriptsize
\setlength{\tabcolsep}{3pt}
\begin{tabular}{@{}lp{5.2cm}lll@{}}
\toprule
Agent & \texttt{\{SCAFFOLD\}} (agent-native) & Blocks & Context & \texttt{\{TOOL\_NOTE\}} \\
\midrule
GPT~5.5 & OpenAI CUA lead-in: tools \texttt{computer}, \texttt{shell}; stop with \texttt{DONE}/\texttt{FAIL}/\texttt{[INFEASIBLE]}. Sent as the first user message. & S,V,P,N,K,G & shell & -- \\
Claude~Opus~4.8 & Claude CUA \texttt{<role>}: tools \texttt{computer}, \texttt{bash}, \texttt{str\_replace\_based\_edit\_tool}; same stop signal. Blocks in \texttt{<instructions>}/\texttt{<safety>}, context in \texttt{<environment>}. & S,V,P,N,K & shell & -- \\
Qwen3.5-35B-A3B & Qwen tool-call prompt: \texttt{computer\_use} schema (1000$\times$1000), XML \texttt{<tool\_call>}, ``Action + tool call'' format, \texttt{terminate(status)}. & -- & shell & bash tool \\
Kimi-K3 & General VLM tool agent (ours): 1280$\times$800 absolute pixels, one tool call per step, \texttt{answer(status)}; tools via function calling. & D & shell & -- \\
OpenCUA-72B & OpenCUA L2 prompt: Thought / Action / Code (PyAutoGUI), \texttt{computer.wait}, \texttt{computer.terminate}. & D & GUI-only & -- \\
EvoCUA-32B & EvoCUA S2 prompt: \texttt{computer\_use} schema (1000$\times$1000), JSON \texttt{<tool\_call>}, ``Action + tool call'' format. & D & GUI-only & -- \\
\bottomrule
\end{tabular}
\end{table}

\subsection{Rubric-Judge Prompt}
\label{app:rubricjudge}
\label{app:llm-judge-prompts}

\begin{lstlisting}[style=prompt,caption={Rubric-judge system prompt.},label={fig:prompt-judgesys}]
You are an expert evaluator of desktop-agent trajectories.

You will receive:
- The user task (for context).
- ONE specific rubric item with a criterion and (optional) verification description.
- The agent's full message/action trajectory (one entry per step), including model
  messages, tool calls, tool outputs, and executed GUI actions.
- Screenshots from the trajectory, in chronological order, as supplementary visual
  evidence for details that the message/action trajectory cannot establish directly.

Your goal is to decide whether this single rubric item is satisfied by the trajectory.

Evaluation rules:
- Judge ONLY the one rubric item you are given; ignore all other implicit requirements.
- Treat the message/action trajectory as the PRIMARY evidence. Read it step by step and
  prioritize recorded model messages, executed actions, tool calls, and tool outputs.
  Use screenshots only as SECONDARY corroborating evidence, especially for visual-only
  state or when the textual trajectory is ambiguous. Do not infer failure merely because
  a completed operation is not clearly visible in a screenshot.
- Evidence counts regardless of its form. A requirement met through a tool call whose
  output appears in the action history is satisfied even when no screenshot depicts it.
  Conversely, a claim that appears ONLY in the agent's final message, with no
  supporting screenshot, action, or tool output, does NOT count.
- Judge the OUTCOME and the route. When a rubric names an application or a UI path
  ("opens the mail app", "in LibreOffice", "navigates to the orders page") but the
  agent reached the same end state by other means -- reading the underlying data
  store, calling an API, generating the file programmatically -- the rubric is
  satisfied, provided the trajectory shows that end state was actually reached.
- Filtering / sorting / form requirements must be applied AND confirmed -- either by
  a screenshot, or by tool output that demonstrates the resulting state.
- If the agent was blocked (captcha, access denied, crash, etc.) and therefore could not satisfy the rubric, report failure.
- If a later step UNDID the rubric (e.g. user-visible state was correct, then was overwritten with wrong data), report failure.

Respond in exactly this format:

Thoughts: <your reasoning, citing specific steps/screenshots/tool output>
Status: "success" or "failure"
\end{lstlisting}

\begin{lstlisting}[style=prompt,caption={Rubric-judge user message.},label={fig:prompt-judgeuser}]
User Task (context only): <TASK_INSTRUCTION>

Primary Message/Action Trajectory:
<ACTION_HISTORY>

Screenshots attached below: <N_SCREENSHOTS> (trajectory had <N_STEPS> total step(s)).

<SCREENSHOT_IMAGE_BLOCKS_IN_CHRONOLOGICAL_ORDER>

Evaluate ONLY this rubric item:
Rubric ID: <RUBRIC_ID>
Requirement: <REQUIREMENT>
Verification: <VERIFICATION>
Weight: <WEIGHT>

Decide whether the rubric (<RUBRIC_ID>) is satisfied. Use the required 'Thoughts:' / 'Status:' format.
\end{lstlisting}

\subsection{Error-Awareness Judge Prompt}
\label{app:judgeprompt}

\begin{lstlisting}[style=prompt,caption={Error-Awareness judge system prompt.},label={fig:judgeprompt}]
You are grading a computer-use agent that took over a task partway through. The steps taken before the takeover contain a mistake, described below by a human annotator. You will read the agent's reasoning segments after the takeover.

Decide whether the agent EXPRESSES RECOGNITION that something went wrong in the earlier steps: for example it notices wrong, missing or fabricated data, a wrong file, page or UI element, that earlier work went off track, or it decides to verify, redo or undo earlier work because it suspects a problem. The recognition does not have to name the exact annotated error, but it must refer to a problem in the work done before the takeover.

Do NOT count: continuing the plan as if nothing is wrong; routine progress narration; generic caution with no reference to a prior problem; a problem that only arises in the agent's own new steps after the takeover.

Answer with strict JSON and nothing else:
{"aware": true or false, "quote": "verbatim excerpt showing the recognition, or an empty string", "reason": "one sentence"}
\end{lstlisting}

\begin{lstlisting}[style=prompt,caption={Error-Awareness judge user message.},label={fig:judgeprompt-user}]
Task instruction:
{instruction}

Annotated error in the steps before the takeover (types: {error_types}):
{error_description}

Agent's reasoning segments after the takeover:

[Thought 1]
{thought_1}

[Thought 2]
{thought_2}

...
\end{lstlisting}

\subsection{Takeover Prompts}
\label{app:historyinjection}

\begin{lstlisting}[style=prompt,caption={Takeover prompt: the first post-prefix user message (after the screenshot) and \texttt{\{takeover\_condition\}} under each condition.},label={fig:takeoverprompt}]
Task: {instruction}
Inspect the current trajectory logs and screenshot and emit the next action as a tool call.

{takeover_condition}
\end{lstlisting}

\begin{lstlisting}[style=prompt]
unaware:
(empty)

notified:
Takeover notice: The preceding trajectory includes one or more serious errors that caused the overall task to fail. Their locations and types are not provided. Identify and fix the errors, then complete the original task.
\end{lstlisting}

\section{Judge Consistency and Human Annotation Protocol}
\label{app:annotation}
\label{app:human-annotation-protocol}

This section describes how humans label failed trajectories and how well our LLM judges agree with human labels.

\subsection{Annotation Protocol}
\label{app:annotator-guidelines}
\label{app:annotation-staffing}
\label{app:double-annotation}

Ten annotators label the failed trajectories. All of them are graduate students in AI.
For each trajectory, the annotator reads the task instruction, the rubric judge rules, the full action trajectories, and the screenshots.
The annotator marks every rubric item as pass or fail.
For a failed trajectory, the annotator also marks the root-cause step, the first step where the failure becomes visible, and the error type.
Each label cites the steps that support it. For Qwen3.5-35B-A3B, EvoCUA-32B, and OpenCUA-72B, we sample $30\%$ of the trajectories, stratified by task category.
A second annotator labels this subset independently.
A third annotator resolves any disagreement.

\subsection{Human--Human Agreement}
\label{app:label-consistency}

Table~\ref{tab:iaa} reports agreement on the Pass@3 judgment of each trajectory generated by Qwen3.5-35B-A3B, EvoCUA-32B, and OpenCUA-72B on \DERAILEVAL.

\begin{table}[h]
\caption{Human--human agreement on the trajectory-level full completion judgment, measured on the $30\%$ double-annotated subsets.}
\label{tab:iaa}
\centering
\small
\begin{tabular}{@{}lccc@{}}
\toprule
Agent & Agree. (\%) & $\kappa$ & F1 \\
\midrule
Qwen3.5-35B-A3B & 97.7 & 0.88 & 0.89 \\
EvoCUA-32B      & 89.2 & 0.77 & 0.85 \\
OpenCUA-72B     & 98.6 & 0.86 & 0.87 \\
\bottomrule
\end{tabular}
\end{table}

\subsection{Human--LLM Agreement}
\label{app:rubricagreement}

The LLM judge decides whether a task passes under Pass@3.
We check it against human judges on \DERAILEVAL.

\paragraph{Setup.}
Human judges labeled $43$ Qwen3.5-35B-A3B tasks and $37$ EvoCUA-32B tasks, $80$ in total.
A task passes under Pass@3 only if any one of three rollouts pass every rubric item.

\paragraph{Results.}
Table~\ref{tab:rubricagree} shows strong agreement: over the $80$ tasks, consistency is $0.938$ and $\kappa$ is $0.818$.
Precision is $1.000$ for both agents, so the LLM judge never passes a task that the human judges fail.
All five disagreements go the other way: the LLM judge fails one Qwen3.5-35B-A3B task and four EvoCUA-32B tasks that the human judges pass.
The LLM judge is therefore slightly strict, and Pass@3 slightly underestimates success, more so on EvoCUA-32B.

\begin{table}[h]
\caption{Agreement between the LLM-as-Judge (GPT~5.6-Terra) and the Human-as-Judge on Pass@3.}
\label{tab:rubricagree}
\centering
\small
\setlength{\tabcolsep}{2.8pt}
\begin{tabular}{@{}lcccccccc@{}}
\toprule
& & \multicolumn{2}{c}{Pass@3} & \multicolumn{5}{c}{LLM-as-Judge vs.\ Human-as-Judge} \\
\cmidrule(lr){3-4}\cmidrule(lr){5-9}
Agent & \#Tasks & Human-as-Judge & LLM-as-Judge & Consistency & Precision & Recall & F1 & $\kappa$ \\
\midrule
Qwen3.5-35B-A3B & 43 & 11.6\% & 9.3\% & 0.977 & 1.000 & 0.800 & 0.889 & 0.876 \\
EvoCUA-32B      & 37 & 40.5\% & 29.7\% & 0.892 & 1.000 & 0.733 & 0.846 & 0.766 \\
\midrule
Total           & 80 & 25.0\% & 18.8\% & 0.938 & 1.000 & 0.750 & 0.857 & 0.818 \\
\bottomrule
\end{tabular}
\end{table}

\section{Experimental Details}
\label{app:baselines}
\label{app:impl}

\subsection{Evaluated Agents}
\label{app:agents}
\label{app:proprietary}
\label{app:opensource}
\label{app:scaffold}

Table~\ref{tab:baselines} lists every agent models with its weights or endpoint and action space. Open-weight models are served with vLLM on AMD MI308X GPUs. APIs-based models are called via their APIs. Every agent keeps the most recent $20$ steps in context. All agents run in the same evaluation harness: the same observation space, snapshots, evaluator, and step budget (3,600 seconds or 150 steps per task). 
\begin{table}[H]
\caption{Evaluated agents. GUI: screenshots with mouse and keyboard actions. CLI: shell commands. Temperature follows each model's reported default.}
\label{tab:baselines}
\label{tab:action-space}
\centering
\footnotesize
\setlength{\tabcolsep}{2.pt}
\resizebox{\linewidth}{!}{\begin{tabular}{@{}lllcc@{}}
\toprule
Agent & Weights / endpoint & Scaffold & GUI & CLI \\
\midrule
Claude~Opus~4.8 & \texttt{claude-opus-4-8} (API) & Claude Code & \checkmark & \checkmark \\
GPT~5.5 & \texttt{gpt-5.5} (API) & Codex & \checkmark & \checkmark \\
Kimi-K3 & \texttt{kimi-k3} (API) & \begin{tabular}[t]{@{}l@{}}general VLM tool agent\\(ours)\end{tabular} & \checkmark & \checkmark \\
Qwen3.5-35B-A3B & \texttt{Qwen/Qwen3.5-35B-A3B} (vLLM) & Qwen CUA & \checkmark & \checkmark \\
EvoCUA-32B & \texttt{meituan/EvoCUA-32B-20260105} (vLLM) & official EvoCUA & \checkmark & -- \\
OpenCUA-72B & \texttt{xlangai/OpenCUA-72B} (vLLM) & official OpenCUA & \checkmark & -- \\
\midrule
\RERAILFULL{} & Qwen3.5-35B-A3B \\ & \WATCHDOG{} (vLLM) & Qwen-CUA & \checkmark & \checkmark \\
\bottomrule
\end{tabular}}
\end{table}

\subsection{SFT Details for Qwen3.5-35B-A3B}
\label{app:impl-train}

We fine-tune Qwen3.5-35B-A3B for one epoch on 64 AMD MI308X GPUs. Table~\ref{tab:train-details} lists the settings. 

\begin{table}[H]
\caption{Training setting of Qwen3.5-35B-A3B.}
\label{tab:train-details}
\centering
\small
\begin{tabular}{@{}ll@{}}
\toprule
Setting & Value \\
\midrule
Base model & Qwen3.5-35B-A3B \\
Training data & \WATCHDOG{}, $8{,}577$ trajectories \\
Hardware & $64 \times$ AMD MI308X, 18 hours \\
Epochs / batch size / learning rate & 1 / 256 / $1\times10^{-5}$, cosine \\
Context length / image history & 32k tokens / 20 screenshots \\
\bottomrule
\end{tabular}
\end{table}

\subsection{Ablation Study}
\label{app:ablation}

To separate recovery-specific supervision from additional training on long in-domain trajectories, we construct a success-only control from successful executions of the same training workflows. We conduct the experiment under the same setting of \WATCHDOG{}. The only difference is the training trajectory: the control follows a successful execution from a clean state, whereas \WATCHDOG{} begins with an erroneous history and demonstrates how to recover from it.

\begin{table}[t]
\caption{Effect of the training trajectory.}
\label{tab:recovery-data-ablation}
\centering
\small
\setlength{\tabcolsep}{4pt}
\resizebox{\linewidth}{!}{%
\begin{tabular}{@{}lrrrrrrr@{}}
\toprule
Training data &
\multicolumn{3}{c}{Recovery from Qwen failures} &
\multicolumn{2}{c}{MyPCBench} &
\multicolumn{2}{c}{Odysseys} \\
\cmidrule(lr){2-4}\cmidrule(lr){5-6}\cmidrule(l){7-8}
& EAR & Rubric Score & Pass@3 & Pass@3 & Rubric Score & Pass@3 & Rubric Score \\
\midrule
None & 19.9 & 5.0 & 0.2 & 9.2 & 33.4 & 24.1 & 66.5 \\
Successful trajectories & 23.6 & 10.7 & 0.3 & 13.8 & 37.4 & 28.4 & 70.2 \\
\WATCHDOG{} recovery trajectories & 35.2 & 16.1 & 1.9 & 24.5 & 53.2 & 34.9 & 78.2 \\
\bottomrule
\end{tabular}%
}
\end{table}

\subsection{Recovery Generation and Supervised Training}
\label{app:recovery-training}

For an erroneous prefix ending at step $c$, we generate continuations using hints of increasing detail. The first hint $\ell_1$ identifies the erroneous step; $\ell_2$ adds the conflicting observation; $\ell_3$ also identifies affected rows. The base model gets up to eight attempts at each level. If these fail, the teacher gets one attempt with $\ell_3$. We stop at the first continuation accepted by the fixed final verifier:
\begin{equation}
r^\star=\operatorname{first}\left\{r\in(r_{1,1},\ldots,r_{1,8},r_{2,1},\ldots,r_{2,8},r_{3,1},\ldots,r_{3,8},r_{\mathrm{T}}):\hat\varphi_{\mathrm{final}}(\xi_{\le c}\oplus r)=1\right\}.
\end{equation}
Here $r_{k,j}$ is generated by the base model conditioned on the task $l$, history $H_c$, and hint $\ell_k$; $r_{\mathrm{T}}$ is generated by the teacher. Prefixes with no verified continuation do not contribute a recovery target. We further check and delete whether $r^\star$ contains information that comes from hints to mitigate the effect of hints on the training.

\subsection{Recovery across Failure Sources}
\label{app:training-transfer}

\begin{figure}[!t]
\centering
\IfFileExists{depth_cross_takeover.pdf}{%
\includegraphics[width=\linewidth]{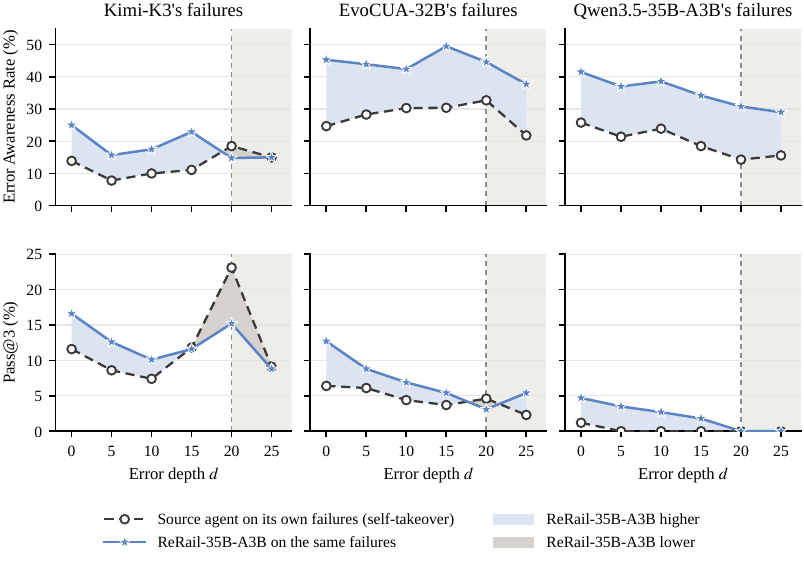}%
}{%
\fbox{\parbox{0.96\linewidth}{\centering\vspace{5em}\textcolor{red}{\textbf{Placeholder: upload horizon\_cross\_takeover.pdf}}\vspace{5em}}}%
}
\caption{\textbf{\RERAILFULL{} recovers other agents' failures better than the agents themselves while the root cause is in context.} Each column fixes one source agent. Top: Error Awareness Rate; bottom: Pass@3. Dashed: self-takeover; blue: \RERAILFULL{} on the same failures (blue fill: higher; gray fill: lower). The shaded region marks $d \ge 20$, where the root-cause screenshot reaches or leaves the 20-screenshot context.}
\label{fig:cross}
\end{figure}

We evaluate \RERAILFULL{} on failure states produced by Qwen3.5-35B-A3B, Kimi-K3, and EvoCUA-32B. Within each pair of rows, the source model and the fine-tuned model start from the same takeover states.Table~\ref{tab:training-awareness} reports EAR on these states, and Figure~\ref{fig:cross} plots EAR and Pass@3 across takeover depths for each failure source.
\begin{table}[ht]
\caption{Error-Awareness Rate (\%) across failure sources and takeover depths. ``Source model'' denotes the agent that produced the failures and is evaluated on its own states.}
\label{tab:training-awareness}
\centering
\small
\setlength{\tabcolsep}{3pt}
\begin{tabular}{@{}llrrrrrr@{}}
\toprule
Failure source & Evaluated agent &
$d{=}0$ & $d{=}5$ & $d{=}10$ & $d{=}15$ & $d{=}20$ & $d{=}25$ \\
\midrule
Qwen3.5-35B-A3B & Source model
& 25.8 & 21.4 & 23.9 & 18.5 & 14.3 & 15.6 \\
& \RERAILFULL
& 41.5 & 37.0 & 38.6 & 34.2 & 30.8 & 29.0 \\
\midrule
Kimi-K3 & Source model
& 13.9 & 7.8 & 10.0 & 11.1 & 18.5 & 15.0 \\
& \RERAILFULL
& 25.0 & 15.7 & 17.5 & 22.9 & 14.8 & 15.0 \\
\midrule
EvoCUA-32B & Source model
& 24.7 & 28.3 & 30.3 & 30.4 & 32.7 & 21.8 \\
& \RERAILFULL
& 45.3 & 43.9 & 42.4 & 49.5 & 44.6 & 37.7 \\
\bottomrule
\end{tabular}
\end{table}

\begin{table}[ht]
\caption{Rubric Score / Pass@3 (\%) for self-takeover, on the same runs as Table~\ref{tab:recovery-main}. Each row uses the named agent\textquotesingle{}s own failures.}
\label{tab:selftakeover}
\label{tab:rerail}
\label{tab:recovery-depth}
\centering
\compactresulttable
\resizebox{0.88\linewidth}{!}{%
\begin{tabular}{@{}lrrrrrr@{}}
\toprule
 & \multicolumn{6}{c}{Rubric Score\,/\,Pass@3 (\%) by takeover depth} \\
\cmidrule(l){2-7}
Agent & $d{=}0$ & $d{=}5$ & $d{=}10$ & $d{=}15$ & $d{=}20$ & $d{=}25$ \\
\midrule
\multicolumn{1}{@{}l}{\emph{Self-takeover}} & & & & & & \\
\multicolumn{1}{@{}l}{\quad\emph{API-based}} & & & & & & \\
GPT~5.5 & 65.3\,/\,20.1 & 67.7\,/\,24.3 & 61.2\,/\,20.9 & 60.8\,/\,20.1 & 68.5\,/\,25.5 & 54.9\,/\,18.9 \\
Claude~Opus~4.8 & 35.3\,/\,0.0 & 24.0\,/\,0.0 & 29.5\,/\,4.5 & 38.5\,/\,5.3 & 48.1\,/\,7.7 & 36.4\,/\,0.0 \\
Kimi-K3 & 39.2\,/\,11.6 & 41.7\,/\,8.6 & 36.4\,/\,7.4 & 33.1\,/\,11.8 & 41.5\,/\,23.1 & 36.9\,/\,9.1 \\
\multicolumn{1}{@{}l}{\quad\emph{Open-source}} & & & & & & \\
EvoCUA-32B & 25.6\,/\,6.4 & 27.1\,/\,6.1 & 23.8\,/\,4.4 & 19.3\,/\,3.7 & 24.5\,/\,4.6 & 16.1\,/\,2.3 \\
Qwen3.5-35B-A3B & 8.4\,/\,1.2 & 6.5\,/\,0.0 & 4.2\,/\,0.0 & 2.1\,/\,0.0 & 4.6\,/\,0.0 & 3.9\,/\,0.0 \\
OpenCUA-72B & 6.0\,/\,0.9 & 4.8\,/\,0.0 & 3.4\,/\,0.0 & 4.7\,/\,0.0 & 3.6\,/\,0.0 & 5.7\,/\,0.0 \\
\bottomrule
\end{tabular}%
}
\end{table}

\clearpage
\section{More Analysis}
\label{app:analysis}

We examine how awareness and recovery change with takeover depth, use recorded failures to explain the difficulties of continuing an erroneous workflow, and discuss how the timing of visible evidence affects the recovery setting.

\subsection{Awareness and Recovery across Takeover Depth}
\label{app:depth-figs}

\begin{figure}[H]
\centering
\IfFileExists{depth_self_takeover.pdf}{%
\includegraphics[width=\linewidth]{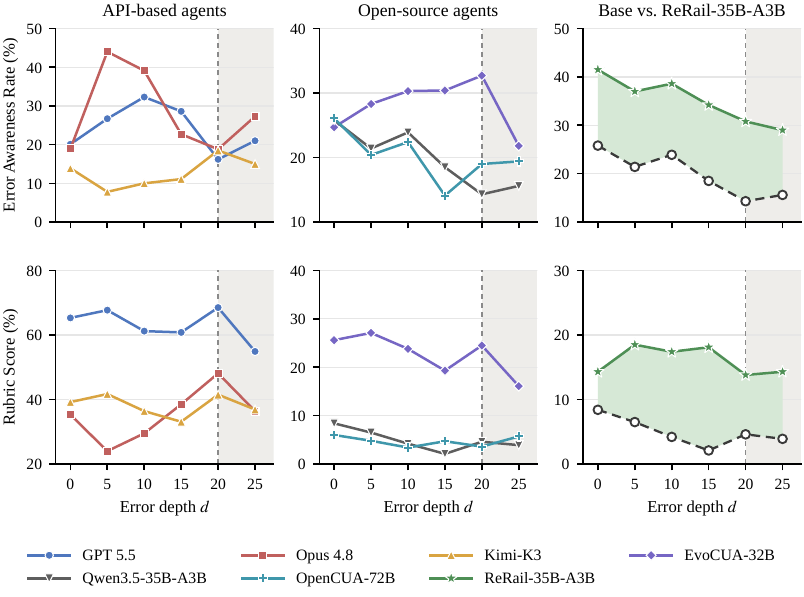}%
}{%
\fbox{\parbox{0.96\linewidth}{\centering\vspace{3.5em}\textcolor{red}{\textbf{Placeholder: upload horizon\_self\_takeover.pdf}}\vspace{3.5em}}}%
}
\caption{\textbf{Awareness and recovery across error depth (self-takeover).} Top: Error Awareness Rate; bottom: Rubric Score. Right column: \RERAILFULL{} against Qwen3.5-35B-A3B on the same failures (green fill: gain; gray fill: loss). The shaded region marks $d \ge 20$, where the root-cause screenshot reaches or leaves the 20-screenshot context. Pass@3 is in Table~\ref{tab:selftakeover}.}
\label{fig:depth}
\end{figure}

\paragraph{Noticing a problem and completing the task follow different patterns.}
The self-takeover results in Tables~\ref{tab:recovery-main} and~\ref{tab:selftakeover} show several points where awareness increases while completion decreases. For EvoCUA-32B, EAR rises from 24.7\% at $d=0$ to 30.4\% at $d=15$, while Pass@3 falls from 6.4\% to 3.7\%. Claude~Opus~4.8 reaches its highest EAR, 44.0\%, at $d=5$, where Pass@3 is 0.0\%. EAR captures whether the agent questions earlier work in its first three reasoning segments. Completing the task also requires finding the relevant mistake, correcting the affected state, and finishing the remaining work. These results make it useful to evaluate early awareness and final completion separately.Figure~\ref{fig:depth} plots EAR (top) and Rubric Score (bottom) across takeover depths for each agent's self-takeover, and compares \RERAILFULL{} with Qwen3.5-35B-A3B on the same failures (right column).

\paragraph{Later takeovers can expose symptoms while losing earlier visual evidence.}
Table~\ref{tab:late-takeover-changes} summarizes the change from $d=20$ to $d=25$. At $d=25$, the root-cause screenshot is outside the 20-screenshot history. All four agents with nonzero Pass@3 at $d=20$ have lower Pass@3 at $d=25$. For GPT~5.5 and Claude~Opus~4.8, EAR increases over the same interval. This pattern is consistent with a difficulty that arises as errors persist through a workflow. Later observations can reveal that something is wrong, while the earlier screenshot needed to understand how it happened is no longer in the supplied history. Recovery then requires the agent to revisit the relevant application or source data and reconstruct the cause from what remains.

\begin{table}[t]
\caption{Changes from $d=20$ to $d=25$ in self-takeover, in percentage points. Each change is the score at $d=25$ minus the score at $d=20$, using Tables~\ref{tab:recovery-main} and~\ref{tab:selftakeover}. Each depth uses its eligible test cases.}
\label{tab:late-takeover-changes}
\centering
\small
\setlength{\tabcolsep}{7pt}
\begin{tabular}{@{}lrrr@{}}
\toprule
Agent & $\Delta$ EAR & $\Delta$ Rubric Score & $\Delta$ Pass@3 \\
\midrule
GPT~5.5 & $+4.8$ & $-13.6$ & $-6.6$ \\
Claude~Opus~4.8 & $+8.5$ & $-11.7$ & $-7.7$ \\
Kimi-K3 & $-3.5$ & $-4.6$ & $-14.0$ \\
EvoCUA-32B & $-10.9$ & $-8.4$ & $-2.3$ \\
Qwen3.5-35B-A3B & $+1.3$ & $-0.7$ & $0.0$ \\
OpenCUA-72B & $+0.4$ & $+2.1$ & $0.0$ \\
\bottomrule
\end{tabular}
\end{table}

\paragraph{Training helps with awareness more consistently than complete recovery.}
On the base Qwen model's failure states, \RERAILFULL has higher EAR and Rubric Score at all six depths (Tables~\ref{tab:training-awareness} and~\ref{tab:recovery-main}). At $d=25$, EAR is 29.0\%, compared with 15.6\% for the base model, and Rubric Score is 14.3\%, compared with 3.9\%. Pass@3 remains 0.0\% for both models. The recovery trajectories help the model notice errors and satisfy more task requirements, but later takeover states remain difficult to complete reliably.

\paragraph{The outcome also depends on the source of the failure.}
On Kimi-K3's states at $d=0$, \RERAILFULL has higher Pass@3 than Kimi-K3 itself, 16.6\% versus 11.6\%, but a lower Rubric Score, 36.4\% versus 39.2\%. This combination holds at $d=5$ and $d=10$ as well. A policy can therefore complete more cases in all three runs while earning less partial credit on average. On EvoCUA-32B's states, the fine-tuned model has higher EAR at every depth, but lower Pass@3 at $d=20$ and $d=25$. Comparing policies on shared states reveals these differences in awareness, partial progress, and reliable completion.

\subsection{Error Horizon and Recovery Outcome}
\label{app:res-horizon}

\paragraph{Evidence timing and takeover depth describe different parts of a failure.}
For a root cause at step $t_r$, the annotated error horizon is $h_e=t_e-t_r$, where $t_e$ is the first step at which the error becomes explicit in the recorded rollout. Takeover depth is $d=c-t_r$, where $c$ is the step at which another agent resumes. Thus, $h_e$ describes when evidence appears along the original execution, while $d$ determines how much of that execution the evaluated agent inherits. The Sweetgreen example has $h_e=40$; taking over at $d=0$ or $d=25$ presents different histories of the same underlying failure.

\paragraph{A long error horizon can separate the cause from its visible consequence.}
For an error with $h_e>20$, a takeover more than 20 steps after the root cause can occur before the error becomes explicit in the original rollout. The supplied screenshot history can then omit the root cause while the current state still looks compatible with the agent's earlier decisions. As the workflow continues, later observations may provide more evidence of the mistake. Recovering also requires connecting that evidence to the earlier decision and checking the data it affected. This offers an explanation for the combination of higher EAR and lower completion seen at some later takeover depths in Section~\ref{sec:depth-analysis}.

\paragraph{Each depth selects a different part of the failure set.}
The construction in Appendix~\ref{app:depthinstantiation} retains takeover points at or before the annotated evidence step and within the recorded rollout. For a failure with an observed $t_e$, these conditions are $d\leq h_e$ and $t_r+d\leq T$, where $T$ is the rollout length. The number of eligible states decreases from 602 at $d=0$ to 231 at $d=25$. Later depths therefore focus on failures that continue long enough to support the takeover and whose explicit evidence appears sufficiently late. Scores at those depths describe recovery in this delayed-evidence setting. The shared-state comparisons in Appendix~\ref{app:training-transfer} hold the inherited state fixed when comparing recovery policies within each depth.

\paragraph{Evidence delay and failure duration need to be read separately.}
The Sweetgreen failure has a long delay before the error becomes explicit. The SprintBoard failure instead contains a long sequence of waits despite visible evidence that the command never ran. Both produce extended erroneous histories, but they call for different recovery behavior. The first requires checking an earlier assumption against source data; the second requires reconsidering an explanation that conflicts with the current screen. Reading the annotated horizon alongside the trajectory and the recovery scores helps distinguish difficulty finding evidence from difficulty acting on it.

\subsection{Recovery Conditioned on Error Awareness}
\label{app:conditional-awareness}

The main results report error awareness and task completion as separate metrics. Here, we examine their relationship directly at the rollout level. For each agent, we divide its self-takeover runs according to whether the agent recognizes the inherited error. We then report the final verifier pass rate within the aware and unaware groups.

\begin{table}[H]
\caption{Recovery conditioned on error awareness in self-takeover runs. Pass denotes acceptance by the final verifier for an individual rollout. Macro Avg. gives equal weight to each agent.}
\label{tab:conditional-awareness}
\centering
\footnotesize
\setlength{\tabcolsep}{5pt}\renewcommand{\arraystretch}{0.92}
\begin{tabular}{lrr}
\toprule
Agent & Pass $|$ Aware & Pass $|$ Unaware \\
\midrule
GPT~5.5 &  34.7 & 31.1 \\
Claude~Opus~4.8 &  13.9 & 7.6 \\
Kimi-K3 & 27.0 &  22.8 \\
EvoCUA-32B & 9.6  & 7.2 \\
Qwen3.5-35B-A3B & 3.2 & 2.8 \\
OpenCUA-72B & 4.9 & 1.5 \\
\midrule
Macro Avg. & 15.6 & 12.2 \\
\bottomrule
\end{tabular}
\end{table}

\end{document}